\documentclass[10pt,twocolumn,letterpaper]{article}

\usepackage{iccv}
\usepackage{times}
\usepackage{epsfig}
\usepackage{graphicx}
\usepackage{amsmath}
\usepackage{amssymb}
\usepackage{bbold}
\usepackage{dsfont}
\usepackage{url}

\usepackage{enumerate}

\usepackage[font=small,labelfont=bf]{caption}

\usepackage{xcolor}
\usepackage{amsthm}
\usepackage{amsfonts}
\usepackage[vlined, ruled]{algorithm2e}
\usepackage{setspace}
\usepackage{caption}
\usepackage{subcaption}
\usepackage{bm}
\usepackage{isomath}
\usepackage{algorithm2e}
\usepackage[numbers]{natbib}
\usepackage[toc,page]{appendix}

\usepackage[pagebackref=true,breaklinks=true,letterpaper=true,colorlinks,bookmarks=false]{hyperref}
\usepackage{footmisc}
\usepackage{multirow}
\usepackage{stmaryrd}
\usepackage{float}
\usepackage{fixltx2e}
\usepackage{dblfloatfix}
\usepackage{pbox}
\usepackage{xcolor}
\usepackage[normalem]{ulem}
\usepackage[toc,page]{appendix}
\usepackage{cleveref}% load last
\usepackage{url}
\usepackage{enumitem}
\usepackage{arydshln}
\usepackage{multirow}
\usepackage{bbding}

\renewcommand\vec[1]{\ensuremath\boldsymbol{#1}}
\renewcommand\cdots{...}

\newcommand{\vb}{\mathbf{b}}
\newcommand{\vy}{\mathbf{y}}

\newcommand{\tX}{\vec{\mathcal{X}}}

\newcommand{\vx}{\mathbf{x}}

\newcommand{\mbr}[1]{\mathbb{R}^{#1}}

\newcommand{\vv}{\mathbf{v}}

\newcommand{\idx}[1]{\mathcal{I}_{#1}}

\newcommand{\vphi}{\boldsymbol{\phi}}
\newcommand{\vpsi}{\boldsymbol{\psi}}

\DeclareMathOperator*{\argmin}{arg\,min}

\DeclareMathOperator*{\avg}{avg}
\DeclareMathOperator*{\sgn}{Sgn}

\newcommand{\expl}[1]{\text{e}^{#1}}

\newtheorem{proposition}{Proposition}
\newtheorem{remark}{Remark}

\newcommand{\vsss}{\boldsymbol{s}}
\newcommand{\vh}{\boldsymbol{h}}

\def\eg{\emph{e.g.}}

\newcommand{\mIdent}{\boldsymbol{\mathds{I}}}

\newcommand{\vOnes}{\mathbb{1}}

\newcommand{\mM}{\boldsymbol{M}}

\newcommand{\mW}{\boldsymbol{W}}

\newcommand{\vm}{\boldsymbol{m}}

\newcommand{\mP}{\boldsymbol{\Theta}}
\newcommand{\mPP}{\boldsymbol{P}}

\newcommand{\stkout}[1]{{\ifmmode\text{\sout{\ensuremath{#1}}}\else\sout{#1}\fi}}

\DeclareMathOperator*{\arcsinh}{arcsinh}
\newcommand{\comment}[1]{}

\makeatletter
\DeclareRobustCommand\onedot{\futurelet\@let@token\bmv@onedotaux}
\def\bmv@onedotaux{\ifx\@let@token.\else.\null\fi\xspace}
\def\eg{\emph{e.g}\onedot}

\def\etc{\emph{etc}\onedot}

\makeatother

\newcommand{\CO}{\color{black!40!blue}}

\usepackage[pagebackref=true,breaklinks=true,letterpaper=true,colorlinks,bookmarks=false]{hyperref}

\iccvfinalcopy % *** Uncomment this line for the final submission

\def\iccvPaperID{2979} % *** Enter the ICCV Paper ID here

\begin{document}

%%%%%%%%% TITLE
%\title{Hallucinating Bag-of-Words and Fisher Vector IDT terms for CNN-based Action Recognition\vspace{-0.3cm}}
%\title{Hallucinating Bag-of-Words and Fisher Vector IDT Terms, and I3D Optical Flow Features for CNN-based Action Recognition\vspace{-0.3cm}}
\title{Hallucinating IDT Descriptors and I3D Optical Flow Features for Action Recognition with CNNs\vspace{-0.35cm}}

\author{Lei Wang\thanks{{\CO Both authors contributed equally. This paper is accepted by the ICCV'19.} Please respect the authors' efforts by not copying/plagiarizing bits and pieces of this work for your own gain (we will vigorously pursue dishonest authors). If you find anything inspiring in this work, be kind enough to cite it thus showing you care for the CV community.
}\textsuperscript{$\;\,$,1,2}\qquad Piotr Koniusz\textsuperscript{$*$,1,2}\qquad Du Q. Huynh\textsuperscript{3}\\
$^1$Data61/CSIRO, $^2$Australian National University, $^3$University of Western Australia\\
firstname.lastname@\{data61.csiro.au\textsuperscript{1}, anu.edu.au\textsuperscript{2}, uwa.edu.au\textsuperscript{3}\}
\vspace{-0.5cm}
}

\maketitle

\begin{abstract}
In this paper, we revive the use of old-fashioned handcrafted video representations for action recognition and put new life into these techniques via a CNN-based hallucination step. %Specifically, we address the problem of action classification in videos via an I3D network pre-trained on the large scale Kinetics-400 dataset.  
Despite of the use of RGB and optical flow frames, the I3D model (amongst others) thrives on combining its output with the Improved Dense Trajectory (IDT) and  extracted with its low-level video descriptors encoded via Bag-of-Words (BoW) and Fisher Vectors (FV). Such a fusion of CNNs and handcrafted representations is time-consuming due to pre-processing, descriptor extraction, encoding and tuning parameters. %fine-tuning of the model. 
Thus, we propose an end-to-end trainable network with streams which learn the IDT-based BoW/FV representations at the training stage and are simple to integrate with the I3D model. 
Specifically, each stream takes I3D feature maps ahead of the last 1D conv. layer and learns to `translate' these maps to BoW/FV representations. Thus, our  model can hallucinate and use such synthesized BoW/FV representations at the testing stage. We show that even features of the entire I3D optical flow stream can be hallucinated thus simplifying the pipeline. 
%
%We demonstrate simplicity of our model and state-of-the-art results on four publicly available datasets.$\!\!$
Our model saves 20--55h of computations and yields state-of-the-art results on four publicly available datasets.$\!\!$
\end{abstract}

%In this paper, we revive the use of old-fashioned handcrafted video representations for action recognition and put new life into these techniques via a CNN-based hallucination step. Despite of the use of RGB and optical flow frames, the I3D model (amongst others) thrives on combining its output with the Improved Dense Trajectory (IDT) and  extracted with its low-level video descriptors encoded via Bag-of-Words (BoW) and Fisher Vectors (FV). Such a fusion of CNNs and handcrafted representations is time-consuming due to pre-processing, descriptor extraction, encoding and tuning parameters. Thus, we propose an end-to-end trainable network with streams which learn the IDT-based BoW/FV representations at the training stage and are simple to integrate with the I3D model. Specifically, each stream takes I3D feature maps ahead of the last 1D conv. layer and learns to `translate' these maps to BoW/FV representations. Thus, our  model can hallucinate and use such synthesized BoW/FV representations at the testing stage. We show that even features of the entire I3D optical flow stream can be hallucinated thus simplifying the pipeline. Our model saves 20-55h of computations and yields state-of-the-art results on four publicly available datasets.

%%%%%%%%% BODY TEXT
\section{Introduction}
\label{sec:intro}

Action Recognition (AR) pipelines have transitioned from the use of handcrafted descriptors \cite{hof,sift_3d,3D-HOG,dense_traj,dense_mot_boundary,improved_traj} to CNN models %\cite{two_stream,spattemp_filters,spat_temp_resnet,i3d_net}. Examples of new powerful CNN architectures are 
such as the two-stream network \cite{two_stream}, 3D spatio-temporal features \cite{spattemp_filters}, spatio-temporal ResNet \cite{spat_temp_resnet} and the I3D network pre-trained on Kinetics-400 \cite{i3d_net}. Such CNNs operate on RGB/optical flow videos thus failing to capture some domain-specific information which sophisticated low-level representations capture by design. One prominent example are Improved Dense Trajectory (IDT) descriptors \cite{improved_traj} which are typically encoded with Bag-of-Words (BoW) \cite{sivic_vq,csurka04_bovw} or Fisher Vectors (FV) \cite{perronnin_fisher,perronnin_fisherimpr} and fused with CNNs \cite{basura_rankpool2,hok,anoop_rankpool_nonlin,anoop_advers,potion} at the classifier %level for the best performance. To understand why CNN-based AR pipelines 
which improves results %upon a fusion with the IDT representations, one has to note that IDT involve 
due to several sophisticated steps of IDT: (i) camera motion estimation, (ii) motion descriptor modeling along motion trajectories estimated by the optical flow, (iii) pruning inconsistent matches, (iv) focusing on human motions
% removing focus from humans 
via a human detector, (v) combination of IDT with powerful and highly complementary to each other video descriptors such as Histogram of Oriented Gradients (HOG) \cite{hog2d,3D-HOG}, Histogram of Optical Flow (HOF) \cite{hof} and Motion Boundary Histogram (MBH) \cite{dense_mot_boundary}   %Moreover, dense trajectories are usually combined with powerful video descriptors such as Histogram  of Oriented Gradients (HOG) \cite{hog2d,3D-HOG}, Histogram of Optical Flow (HOF) \cite{hof} and Motion Boundary Histogram (MBH) \cite{dense_mot_boundary} %, or even SIFT3D \cite{sift_3d} 
%which are highly complementary to each other 
\eg, HOF and MBH  contain zero- and first-order motion statistics \cite{improved_traj}.

\begin{figure*}[t]%htbp % left bottom right top
\centering%%%%
\vspace{-0.3cm}
\centering\includegraphics[trim=0 0 0 1, clip=true,width=\textwidth]{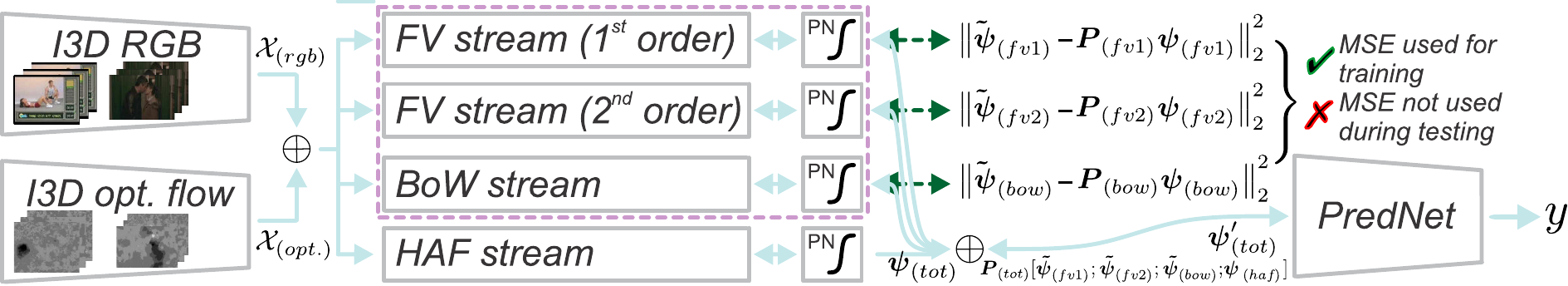}
\caption{The overview of our pipeline. We remove the prediction and the last 1D conv. layers from I3D RGB and optical flow streams, concatenate ($\oplus$) the $1024\!\times\!7$ feature representations $\mathcal{X}_{(rgb)}$ and $\mathcal{X}_{(opt.)}$, and feed them into our {\em Fisher Vector} ({\em FV}), {\em Bag-of-Words} ({\em BoW}), and {\em the High Abstraction Features} ({\em HAF}) streams followed by the {\em Power Normalization} ({\em PN}) blocks. The resulting feature vectors $\vec{\tilde{\psi}}_{(fv1)}$, $\vec{\tilde{\psi}}_{(fv2)}$, $\vec{\tilde{\psi}}_{(bow)}$ and $\vec{\psi}_{(haf)}$ are concatenated ($\oplus$) and fed into our {\em Prediction Network} ({\em PredNet}). By \Checkmark, we indicate that the three Mean Square Error (MSE) losses are only applied at the training stage to train our FV (first- and second-order components) and BoW hallucinating streams (indicated in dashed red). By \XSolidBrush, we indicate that the MSE losses are switched off at the testing stage. Thus, we hallucinate $\vec{\tilde{\psi}}_{(fv1)}$, $\vec{\tilde{\psi}}_{(fv2)}$ and $\vec{\tilde{\psi}}_{(bow)}$, and pass them to PredNet together with $\vec{\psi}_{(haf)}$ to obtain labels $y$. The original training FV and BoW feature vectors (used only during training) are denoted by $\vpsi_{(fv1)}$, $\vpsi_{(fv2)}$ and $\vpsi_{(bow)}$, while $\mPP$ are count sketch projecting matrices (see text for details).
}
\vspace{-0.3cm}
\label{fig:pipeline}
\end{figure*}

However, extracting dense trajectories and corresponding video descriptors is costly due to several off-line/CPU-based steps. %, due to  time-consuming steps described above. 
Motivated by this shortcoming, we propose  simple trainable CNN streams on top of a CNN network (in our case I3D \cite{i3d_net}) which  learn to `translate' the I3D output into IDT-based BoW and FV global video descriptors. We can even `translate' the I3D RGB output into I3D Optical Flow Features (OFF). At the testing stage, our so-called BoW, and FV and OFF streams (on top of I3D) are able to hallucinate such global descriptors which we feed  into the final layer preceding a classifier. We show that %sophisticated handcrafted representations (
IDT/OFF representations can be synthesized by our network thus removing the need of %actual computations of IDT/OFF representations
actually computing them 
 which simplifies the AR pipeline. %Depending on an employed variant of our streams, 
With a handful of convolutional/FC layers and basic CNN building blocks, our representation rivals sophisticated AR pipelines that aggregate features frame-by-frame \eg, HOK \cite{hok} and rank-pooling \cite{basura_rankpool2,anoop_rankpool_nonlin,anoop_advers,potion}. 
Below, we detail our contributions:
\renewcommand{\labelenumi}{\Roman{enumi}.}
\vspace{-1.8mm}
\hspace{-1cm}
\begin{enumerate}[leftmargin=0.6cm]
    \item We are the first to propose that old-fashioned IDT-based BoW and FV global video descriptors can be learned via simple dedicated CNN-streams at the training stage and simply hallucinated for classification with a CNN action recognition pipeline during testing.\vspace{-0.15cm}
		\item We show that even the I3D optical flow stream can be easily hallucinated from the I3D RGB stream.\vspace{-0.15cm}
    \item We study various aspects of our model \eg, the count sketch \cite{pham_sketch} of features to avoid overfitting when fusing several streams and Power Normalization \cite{me_ATN,me_tensor,me_deeper} to prevent so-called burstiness in BoW, FV and CNNs, and we % by removing various blocks of our network. %some streams whilst retaining others.
		%\vspace{-0.2cm}
    %\item 
		 perform several experiments on four datasets. %standard publicly available datasets. % video recognition benchmarks. % thus demonstrating the usefulness of our idea.
		\vspace{-0.15cm}
\end{enumerate}

%We show that even features of the entire I3D optical flow stream can be hallucinated thus simplifying the pipeline. 

%Sections \ref{sec:related} and \ref{sec:backgr} introduce the background, notations and concepts. Sections \ref{sec:approach} and \ref{sec:exper} present our method and results.

%Sections \ref{sec:related} and \ref{sec:backgr} introduce background works and their relation to our method  followed by notations and necessary concepts. Sections \ref{sec:approach} and \ref{sec:exper} outline our approach and results.

%The reminder of this paper is organized as follows. Section \ref{sec:related} discusses background works and their relation to our approach. Section \ref{sec:backgr} introduces notations and necessary concepts. Section \ref{sec:approach} outlines our approach. Finally, Sections \ref{sec:exper} and \ref{sec:concl}  present our empirical results and conclusions.

\section{Related Work}
\label{sec:related}

%In what follows, we describe related works such as early handcrafted spatio-temporal representations of videos and their encoding strategies, optical flow representations, and deep learning pipelines dedicated to video classification followed by related aggregation strategies.

Below, we describe handcrafted spatio-temporal video descriptors and their encoding strategies, optical flow, and deep learning pipelines for video classification. % followed by related aggregation strategies.

\vspace{0.05cm}
\noindent{\bf{Handcrafted video representations.}} Early AR %action recognition approaches 
relied on spatio-temporal interest point detectors \cite{harris3d,cuboid,sstip,hes-stip,mv-stip,dense_traj} and spatio-temporal descriptors \cite{hof,sift_3d,hof2, dense_traj,dense_mot_boundary,improved_traj} which %were designed to 
capture various appearance and motion statistics.

Spatio-temporal interest point detectors were developed for the task of identifying spatio-temporal regions of videos rich in motion patterns relevant to classification, thus providing sampling locations for local descriptors. The number of sampling points had a significant influence on the processing speed due to the volumetric nature of videos. Harris3D \cite{harris3d}, one of the earliest detectors, performs a search for extreme points in the spatio-temporal domain via the so-called structure tensor and the determinant-to-trace ratio test. Cuboid \cite{cuboid}, a faster detector, applies Gaussian and Gabor filters in spatial and temporal domains, respectively. Selective STIP \cite{sstip} extracts initial key-point candidates with the Harris corner detector followed by the candidate suppression with a so-called surround suppression mask. Hes-STIP, a more recent detector, uses integral videos and Hessian matrix to search the scale-space for local maxima of the signal. Evaluations and further reading on spatio-temporal detectors can be found in surveys \cite{key3d_survey, lei_thesis_2017, lei_tip_2019}.

One drawback of spatio-temporal interest point detectors is the sparsity of key-points and inability to capture long-term motion patterns. Thus, a Dense Trajectory (DT) \cite{dense_traj} approach densely samples feature points in each frame to track them in the video (via optical flow). Then, multiple descriptors are extracted along trajectories to capture shape, appearance and motion cues. As DT cannot compensate for the camera motion, the IDT \cite{improved_traj,dense_mot_boundary}  estimates the camera motion to remove the global background motion. IDT also removes inconsistent matches via a human detector. 

For spatio-temporal descriptors, IDT employs HOG \cite{hog2d}, HOF \cite{hof} and MBH \cite{dense_mot_boundary}. 
HOG \cite{hog2d} contains statistics of the amplitude of image gradients w.r.t. the gradient orientation. Thus, it captures the static appearance cues while its close cousin, HOG-3D \cite{3D-HOG}, is designed for spatio-temporal interest points.
In contrast, HOF \cite{hof}  captures histograms of optical flow while MBH \cite{dense_mot_boundary} captures derivatives of the optical flow, thus it is highly resilient to the global camera motion whose cues cancel out due to derivatives. Thus, HOF and MBH contain the zero- and first-order optical flow statistics. Other spatio-temporal descriptors include SIFT3D \cite{sift_3d}, SURF3D \cite{hes-stip} and %Local Trinary Patterns \cite{LTP}, a cousin of well-known LBPs.
LTP \cite{LTP}.

\begin{figure}[b]%htbp % left bottom right top
\centering%%%%
%\hspace{-0.1cm}
\vspace{-0.3cm}
\centering\includegraphics[trim=0 0 0 1, clip=true,width=8.4cm]{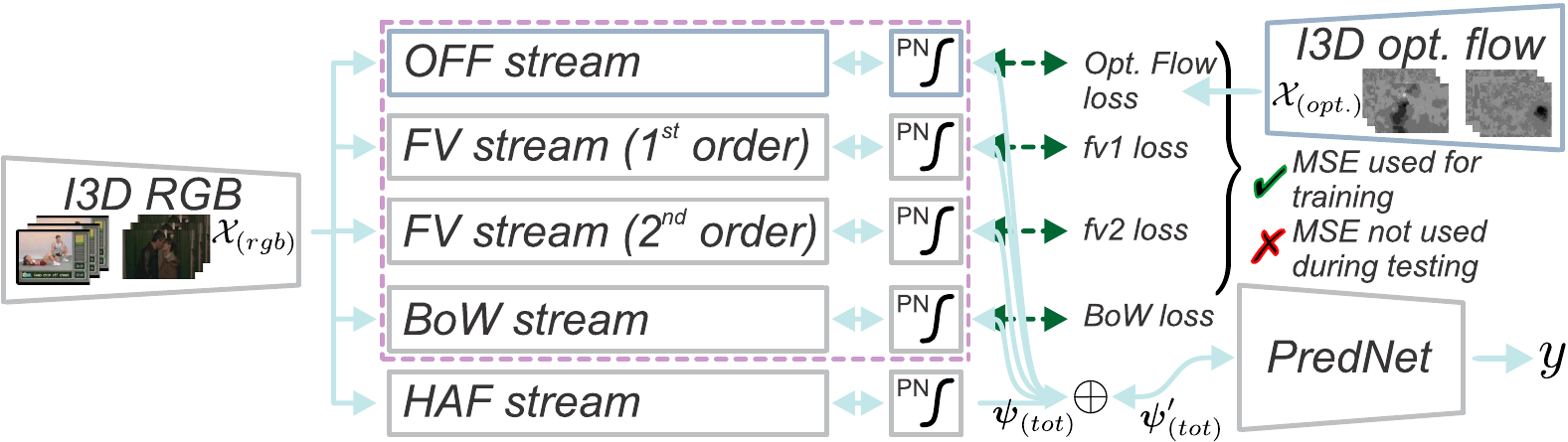}
\caption{Hallucinating the Optical Flow Features (OFF).}
\vspace{-0.3cm}
\label{fig:pipeline2}
\end{figure}

In this work, we follow the standard practice, that is, we use the Improved Dense Trajectories  \cite{dense_traj,hok,potion} and we encode them together with HOG, HOF, and MBH descriptors via BoW \cite{sivic_vq,csurka04_bovw} and FV \cite{perronnin_fisher,perronnin_fisherimpr} which we describe below.

\vspace{0.05cm}
\noindent{\bf{Descriptor encoding.}} BoW \cite{sivic_vq,csurka04_bovw}, a global image representation, %inspired by  natural language processing, 
is likely the oldest encoding strategy for local descriptors. It consists of (i) clustering with k-means for a collection of descriptor vectors from the training set to build so-called visual vocabulary, (ii) assigning each descriptor to its nearest cluster center from the visual dictionary, and (iii) aggregating the one-hot assignment vectors via  average pooling. Similar  models such as Soft Assignment (SA) \cite{soft_ass,me_SAO} and Localized Soft Assignment (LcSA) \cite{liu_sadefense,me_ATN} use the Component Membership Probability (CMP) of GMM to assign each descriptor with some probability to %each/k-closest 
visual words followed by average or non-linear pooling \cite{me_ATN, lei_icip_2019}. %In LcSA, each descriptor is soft-assigned to its k-nearest neighbors.

In this paper, we chose the simplest BoW model \cite{csurka04_bovw} with Power Normalization \cite{me_ATN} detailed in Section \ref{sec:backgr}. BoW can be seen as zero-order statistics of FV \cite{perronnin_fisher,perronnin_fisherimpr}, thus we also employ FV to capture first- and second-order statistics of local descriptors. FV builds a visual dictionary from training data via GMM. Then, a displacement/square displacement of each descriptor vector w.r.t. each GMM component center is taken, normalized by its GMM standard deviation/variance to capture the first/second-order terms, and then soft-assigned via CMP to each GMM component. %Thus, FV are of $2K\!\times\!D$ dimensionality, where $K$ is the number of GMM components and $D$ is the descriptor size ($D$ is often the size after applying  PCA to descriptors).

\vspace{0.05cm}
\noindent{\bf{Optical flow.}}  As a key concept in AR from videos, optical flow is the distribution of velocities of movement of brightness pattern across frames \cite{flow_def2} such as the pattern of motion of objects, surfaces and edges in a visual scene caused by the relative motion between an observer and a scene \cite{flow_def1}. Early optical flow coped with small displacements via energy minimization \cite{flow_def2,brox_accurate}. %and, by surface warping, it could be very accurate  \cite{brox_accurate}. 
However, to capture informative motions of subjects/objects, optical flow needs to cope with large displacements \cite{alvarez_large}. As energy-based methods suffer from the local minima, local descriptor matching is used in Large Displacement Optical Flow (LDOF) \cite{brox_largedisp}. Recent methods use non-rigid descriptor matching \cite{deep_flow}, segment matching \cite{seg_flow} or even edge-preserving interpolation \cite{epic_flow}.%on a sparse set of matches to deal with occlusions \cite{epic_flow}.

In this work, we are not concerned with the use of the newest possible optical flow. Thus, we opt for LDOF \cite{brox_accurate}.

\vspace{0.05cm}
\noindent{\bf{CNN-based action recognition.}} %Despite the great success of handcrafted representations, 
The success of AlexNet \cite{krizhevsky_alexnet} and ImageNet \cite{ILSVRC15} sparked studies into AR with CNNs. Early models extracted per-frame representations followed by average pooling %defined as an average over feature vectors 
\cite{cnn_basic_ar} which discards the temporal order. To fix such a shortcoming, frame-wise CNN scores were fed to LSTMs \cite{cnn_lstm_ar}. Two-stream networks \cite{two_stream} compute representations per RGB frame and per 10 stacked optical flow frames. However, a more obvious extension is to model spatio-temporal 3D CNN filters \cite{cnn3d_ar,spattemp_filters,spat_temp_resnet,long_term_ar}.

%The above pipelines have various merits and drawbacks, thus 
The recent I3D model \cite{i3d_net} draws on the two-stream networks, `inflates' 2D CNN filters pre-trained on ImageNet to spatio-temporal 3D filters, and implements temporal pooling across the inception module. In this paper, we opt for the I3D network but our proposed layers are independent of the CNN design. We are concerned with `absorbing' the old yet powerful IDT representations and/or optical flow features into CNN and hallucinating them at the test time.

\vspace{0.05cm}
\noindent{\bf{Temporal aggregation.}} While two-stream networks \cite{two_stream} discard the temporal order and others use LSTMs \cite{cnn_lstm_ar}, many AR pipelines address the spatio-temporal aggregation. Rank pooling  \cite{basura_rankpool,basura_rankpool2}  projects frame-wise feature vectors onto a line such that the temporal order of vectors is preserved along the line. %Further extensions include 
Subspace and kernel rank pooling \cite{anoop_rankpool_nonlin,anoop_advers} %for which frame-wise representations are projected in 
use projections into the RKHS in which the temporal order of frames is preserved. Another aggregation family captures second- or higher-order statistics \cite{hok,me_tensor_eccv16,me_tensor,Pengfei_ICCV19}.

In this paper, we are not concerned with temporal pooling. Thus, we use a 1D convolution (as in  I3D \cite{i3d_net}). % to factor out the temporal dimension of feature maps.

\vspace{0.05cm}
\noindent{\textbf{Power Normalization family.}} BoW, FV and even CNN-based descriptors have to deal with the so-called burstiness defined as `{\em the property that a given visual element appears more times in an image than a statistically independent model would predict}' \cite{jegou_bursts}, a phenomenon also present in video descriptors. Power Normalization~\cite{me_ATN,me_tensor_tech_rep} is known to suppress the burstiness, and it has been extensively studied in the context of BoW \cite{me_ATN,me_tensor_tech_rep,me_tensor,me_deeper}. Moreover, a connection to Max-pooling was found in survey \cite{me_ATN} which also shows that the so-called MaxExp pooling is in fact a detector of `\emph{at least one particular visual word being present in an image}'. According to papers \cite{me_ATN,me_deeper}, many Power Normalization functions are closely related. We outline Power Normalizations used in our work in Section \ref{sec:backgr}. %Below, we describe the mathematical prerequisites on which we build. 

\section{Background}
\label{sec:backgr}

In our work, we use BoW/FV (training stage), as well as Power Normalization \cite{me_ATN,me_tensor} and count sketches \cite{weinberger_sketch}.

\vspace{+0.05cm}
\noindent{\textbf{Notations.}} We use boldface uppercase letters to express matrices \eg, $\mM, \mPP$, regular uppercase letters with a subscript to express matrix elements \eg, $P_{ij}$ is the $(i,j)^{\text{th}}$ element of $\mPP$, boldface lowercase letters to express vectors, \eg $\vx, \vphi, \vpsi$, and regular lowercase letters to denote scalars. Vectors can be numbered \eg, $\vm_1,\cdots,\vm_K$ or $\vx_n$, \etc, while regular lowercase  letters with a subscript express an element of vector \eg, $m_i$ is the $i^{\text{th}}$ element of $\vm$. Operators `$;$' and $\oplus$ concatenate vectors \eg, $\oplus_{i\in\idx{K}}\vv_i\!=[\vv_1; \cdots; \vv_K]$ while $\idx{d}$ denotes an index set of integers $\{1,\cdots,d\}$.

\subsection{Descriptor Encoding Schemes}
\label{sec:enc}

\vspace{+0.05cm}
\noindent{\textbf{Bag-of-Words}} \cite{sivic_vq,csurka04_bovw} assigns each local descriptor $\vec{x}$ to the closest visual word from $\vec{M}\!=\!\left[\vec{m}_1,\cdots,\vec{m}_K\right]$ built via k-means. In order to  obtain mid-level feature $\vec{\phi}$, we solve:
%The following problem is solved:
%
\begin{equation}\label{eq:sp1}
\begin{array}{l}
\vec{\phi}=\argmin\limits_{\vec{\phi'}}\;\bigr \lVert{ \vec{x}-\vec{M}\vec{\phi'} }\bigr \rVert_2^2,\\
s.\;t.\;\;\vec{\phi'}\in\{0,1\}, \vOnes^T\vec{\phi'}\!=\!1.
\end{array}
\end{equation}
%A low number of non-zero coefficients in $\vec{\phi}$, referred to as sparsity, is induced with the $\ell_1$ norm 
%over $\vec{\phi}$
%and adjusted by constant $\alpha$. We impose a non-negative constraint on $\vec{\phi}$ for compatibility with Analytical pooling~\cite{boureau_pooling, me_ATN}.

\vspace{0.05cm}
\noindent{\textbf{Fisher Vector Encoding}} \cite{perronnin_fisher,perronnin_fisherimpr} uses a Mixture of $K$ Gaussians from a GMM used as a dictionary. It performs descriptor coding w.r.t. to Gaussian components $G(w_k, \vec{m}_k, \vec{\sigma}_k)$ which are parametrized by mixing probability, mean, and on-diagonal standard deviation. The first- and second-order features $\vec{\phi}_k, \vec{\phi}'_k\in\mathds{R}^D$ are :
\begin{equation}\label{eq:fisher1}
%\vec{\phi}_k=\frac{\vec{x}\!-\!\vec{m}_k}{\vec{\sigma}_{k}},\;\;
\vec{\phi}_k=(\vec{x}\!-\!\vec{m}_k)/\vec{\sigma}_{k},\;\;
%
%\vec{\phi}^{(2)}_k=\frac{\left(\vec{x}\!-\!\vec{m}_k\right)^2}{\vec{\sigma}^2_{k}}\!-\!1
%\vec{\phi}^{(2)}_k=\left(\frac{\vec{x}\!-\!\vec{m}_k}{\vec{\sigma}_{k}}\right)^2\!-\!1
\vec{\phi}'_k=\vec{\phi}_k^2\!-\!1.
\end{equation}
Concatenation of per-cluster features $\vec{\phi}^{*}_k\in\mathds{R}^{2D}$ forms the mid-level feature $\vec{\phi}\in\mathds{R}^{2KD}$:
\vspace{-0.2cm}
\begin{equation}\label{eq:fisher2}
\!\!\!\!\vec{\phi}=\left[\vec{\phi}_1^{*}; ...; \vec{\phi}_K^{*} \right],\;\;  \vec{\phi}^{*}_k=\frac{p\left(\vec{m}_k|\vec{x}, \theta\right)}{\sqrt{w_k}}
%
%\begin{bmatrix}
%\vec{\phi}_k\\
%\vec{\phi}'_k/\sqrt{2}
%\end{bmatrix}
\left[\vec{\phi}_k; \vec{\phi}'_k/\sqrt{2}\right],
\!\!\!
\end{equation}
where $p$ and $\theta$ are the component membership probabilities and parameters of GMM, respectively. For each descriptor $\vec{x}$ of dimensionality $D$ (after PCA), its encoding $\vec{\phi}$ is of $2KD$ dim. as $\vec{\phi}$ contains first- and second-order statistics.

\subsection{Pooling a.k.a. Aggregation}
\label{sec:aggr}
%\vspace{0.05cm}
%\noindent{\textbf{Pooling/aggregation. }}

Traditionally, pooling is performed via averaging mid-level feature vectors $\vphi(\vx)$ corresponding to (local) descriptors $\vx\!\in\!\tX$ from a video sequence $\tX$, that is $\vpsi\!=\!\avg_{\vx\in\tX}\vphi(\vx)$, and (optionally) applying the $\ell_2$-norm normalization. In this paper, we work with either sequences $\tX$ (for which the above step is used) or subsequences.
\begin{proposition}
\label{pr:subseq}
For subsequence pooling, let $\tX_{s,t}\!=\!\tX_{0,t}\!\setminus\tX_{0,s-1}$, where $\tX_{s,t}$ denotes a set of descriptors in the sequence $\tX$ counting from frame $s$ up to frame $t$, where $0\!\leq\!s\!\leq\!t\!\leq\tau$, $\tX_{0,-1}\!\equiv\!\emptyset$, and $\tau$ is the length of $\tX$. Moreover, let us compute an integral mid-level feature $\vphi'_t\!=\!\vphi'_{t-1}\!+\!\sum_{\vx\in\tX_{t,t}}\vphi(\vx)$ which aggregates mid-level feature vectors from frame $0$ to frame $t$, and $\vphi'_{-1}$ is an all-zeros vector. Then, the pooled subsequence is given by:
\begin{equation}\label{eq:integr1}
%\vpsi_{s,t}\!=\frac{\vphi'_t\!-\!\vphi'_{s-1}}{\lVert\vphi'_t\!-\!\vphi'_{s-1}\rVert_2}.
\vpsi_{s,t}\!={(\vphi'_t\!-\!\vphi'_{s-1})}/({\lVert\vphi'_t\!-\!\vphi'_{s-1}\rVert_2}+\epsilon),
\end{equation}
where $0\!\leq\!s\!\leq\!t\!\leq\tau$ are the starting and ending frames of subsequence $\tX'_{s,t}\!\subseteq\!\tX$ and $\epsilon$ is a small constant. We normalize the pooled sequences/subseq. as described next.
\end{proposition}

\subsection{Power Normalization}
\label{sec:pns}
%\vspace{0.05cm}
%\noindent{\textbf{Power Normalization.}} 
As alluded to in Section \ref{sec:related}, we apply Power Normalizing functions to BoW and FV streams which hallucinate these two modalities (and HAF/OFF stream explained later). We investigate three operators $g(\vpsi,\cdot)$  detailed by Remarks \ref{re:asinhe}--\ref{re:axmin}.

\begin{remark}
\label{re:asinhe}
AsinhE function \cite{me_deeper} is an extension of a well-known Power Normalization (Gamma) \cite{me_deeper} defined as $g(\vpsi, \gamma)\!=\!\sgn(\vpsi)|\vpsi|^\gamma$ for $0\!<\!\gamma\!\leq\!1$ to the operator with a smooth derivative and a parameter $\gamma'$. AsinhE is defined as the normalized Arcsin hyperbolic function:
\begin{align}
& \!\!\!\!\!g(\vpsi, \gamma')\!=\arcsinh(\gamma'\!\vpsi)/\arcsinh(\gamma'). %\!=\!\log(\gamma'\!\vpsi+\sqrt{1+{\gamma'}^2\!\vpsi^2}).
\end{align}
\end{remark}

\begin{remark}
\label{re:sigme}
Sigmoid (SigmE), a Max-pooling approximation \cite{me_deeper}, is an extension of the MaxExp operator defined as $g(\vpsi, \eta)\!=\!1\!-\!(1\!-\!\vpsi)^{\eta}$ for $\eta\!>\!1$ to the operator with a smooth derivative, a response defined for real-valued $\vpsi$ (rather than $\vpsi\!\geq\!0$), a parameter $\eta'$ and a small const. $\epsilon'$:
\begin{align}
& \!\!\!\!\!g(\vpsi, \eta')\!=\!\frac{2}{1\!+\!\expl{{-\eta'\vpsi}/{(\lVert\vpsi\rVert_2+\epsilon')}}}\!-\!1.
\label{eq:sigmoid}
\end{align}
\end{remark}

\begin{remark}
\label{re:axmin}
AxMin, a piece-wise linear form of SigmE \cite{me_deeper}, is given as $g(\vpsi, \eta'')\!=\!\sgn(\vpsi)\min(\eta''\vpsi/(\lVert\vpsi\rVert_2+\epsilon'), 1)$ for $\eta''\!>\!1$ and a small constant $\epsilon'$.
\end{remark}

Despite the similar role of these three pooling operators, we investigate each of them as their interplay with end-to-end learning differs. Specifically, $\lim_{\vpsi\rightarrow\pm\infty}g(\vpsi,\cdot)$ for AsinhE and SigmE are $\pm\!\infty$ and $\pm\!1$, resp., thus their asymptotic behavior differs. Moreover, AxMin is non-smooth and relies on the same gradient re-projection properties as ReLU.

\subsection{Count Sketches}
\label{sec:sketch}

%Feature maps corresponding to polynomial kernels can be sketched for rapid dimensionality reduction via so-called tensor sketching \cite{pham_sketch}. While tensor-sketching requires FFT and IFFT to combine sketches in each mode due to outer-product operations, 
Sketching vectors by the count sketch \cite{cormode_sketch,weinberger_sketch} is used for their dimensionality reduction which we use in this paper.
\begin{proposition}
\label{pr:ten_sketch}
Let $d$ and $d'$ denote the dimensionality of the input and sketched output vectors, respectively. Let vector $\vh\!\in\!\idx{d'}^d$ contain $d$ uniformly drawn integer numbers from $\{1,\cdots,d'\}$ and vector $\vsss\!\in\!\{-1,1\}^{d}$ contain $d$ uniformly drawn values from $\{-1,1\}$. Then, the sketch projection matrix $\mPP\!\in\!\{-1,0,1\}^{d'\times d}$ becomes:
\begin{equation}\label{eq:sk1}
P_{ij}\!=\!
\begin{cases} s_i  & \text{if }h_i\!=\!j,
\\
0 &\text{otherwise},
\end{cases}%\vspace{-0.05cm}
\end{equation}
and the sketch projection $p: \mbr{d}\!\rightarrow\!\mbr{d'}$  is a linear operation given as $p(\vpsi)\!=\!\mPP\vpsi$ (or $p(\vpsi; \mPP)\!=\!\mPP\vpsi$ to highlight $\mPP$).
\begin{proof}
\vspace{-0.15cm}
It directly follows from the definition of the count sketch \eg, see Definition 1 \cite{weinberger_sketch}.
\end{proof}
\end{proposition}

\begin{remark}
\label{re:ten_sketch}
Count sketches are unbiased estimators: $\mathbb{E}_{\vh,\vsss}(p(\vpsi,\mPP(\vh,\vsss)),p(\vpsi',\mPP(\vh,\vsss)))\!=\!\left<\vpsi,\vpsi'\right>$. As variance $\mathbb{V}_{\vh,\vsss}(p(\vpsi),p(\vpsi'))\!\leq\!\frac{1}{d'}\left(\left<\vpsi,\vpsi'\right>^2 +\lVert\vpsi\rVert_2^2\lVert\vpsi'\rVert_2^2\right)$, we %$\ell_2$-norm normalize input feature vectors.
note that larger sketches are less noisy. Thus, for every modality we compress, we use a separate sketch matrix $\mPP$. As video modalities are partially dependent, this implicitly leverages the unbiased estimator and reduces the variance.
\begin{proof}
\vspace{-0.15cm}
For the first and second property, see  Appendix A of paper \cite{weinberger_sketch} and Lemma 3 \cite{pham_sketch}.
\end{proof}
\end{remark}

\begin{figure}[t]%htbp % left bottom right top
\centering%%%%
\vspace{-0.3cm}
\begin{subfigure}[b]{0.98\linewidth}
\centering\includegraphics[trim=0 0 0 0, clip=true,width=0.98\linewidth]{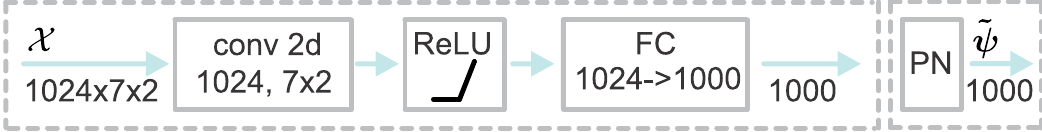}
\vspace{-0.2cm}
\caption{\label{fig:stra}}
\vspace{0.2cm}
\end{subfigure}
\\
\begin{subfigure}[b]{0.98\linewidth}
\centering\includegraphics[trim=0 0 0 0, clip=true,width=0.98\linewidth]{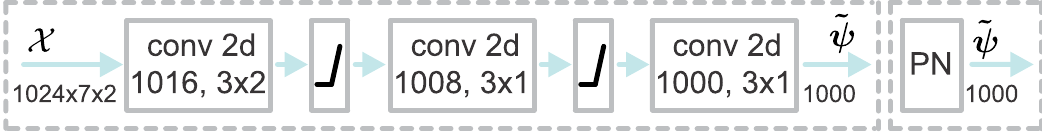}
\vspace{-0.2cm}
\caption{\label{fig:strb}}
\vspace{0.2cm}
\end{subfigure}
\\
\begin{subfigure}[b]{0.98\linewidth}
\centering\includegraphics[trim=0 0 0 0, clip=true,width=0.98\linewidth]{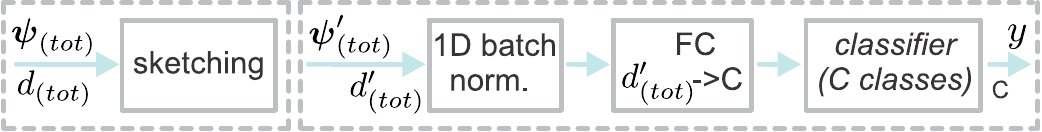}
\vspace{-0.2cm}
\caption{\label{fig:strc}}
\vspace{-0.2cm}
\end{subfigure}
%
%\vspace{-0.7cm}
\caption{Stream types used in our network. Figures \ref{fig:stra} and \ref{fig:strb} show {\em Fully Connected} and {\em Convolutional} variants used for the practical realization of the FV, BoW, OFF and HAF streams. Figure \ref{fig:strc} shows our PredNet. Note that we indicate the type of operation and its parameters in each block \eg, {\em conv2d} and its number of filters/size, or {\em Power Normalization} ({\em PN}). Beneath arrows, we  indicate the size of input, intermediate or output representation.
}\vspace{-0.3cm}
\label{fig:streams}
\end{figure}

\vspace{-0.1cm}
\section{Approach}
\label{sec:approach}

%With all related mathematical concepts described above, we can proceed to defining our AR pipeline below.

Our pipeline is illustrated in Figure \ref{fig:pipeline}. It consist of (i) the Fisher Vector and Bag-of-Words hallucinating streams denoted as FV and BoW (shown in dashed red), respectively, (ii) the High Abstraction Features stream denoted as HAF, and (iii) the Prediction Network abbreviated as PredNet.

The role of BoW/FV streams is to take I3D intermediate representations generated from the RGB and optical flow frames and learn to hallucinate BoW/FV representations. For this purpose, we use the MSE loss between the ground-truth BoW/FV and the outputs of BoW/FV streams. The role of the HAF stream is to further process I3D intermediate representations before they are concatenated with hallucinated BoW/FV. PredNet fuses the concatenated BoW/FV/HAF  and learns class concepts. % of the corresponding actions. 
Figure \ref{fig:pipeline2} shows our pipeline for hallucinating the OFF representation (I3D optical flow). % we take I3D intermediate representations generated from the RGB frames while I3D optical flow is precomputed.
Below, we describe each module in detail.

\subsection{BoW/FV Hallucinating Streams}
\label{sec:fv-bow}

BoW/FV take as input the I3D intermediate representations $\mathcal{X}_{(rgb)}$ and $\mathcal{X}_{(opt.)}$  of size $1024\!\times\!7$ which were obtained by stripping the classifier and the last 1D conv. layer of I3D pre-trained on Kinetics-400. The latter dimension of $\mathcal{X}_{(rgb)}$ and $\mathcal{X}_{(opt.)}$ can be thought of as the temporal size. We concatenate $\mathcal{X}_{(rgb)}$ and $\mathcal{X}_{(opt.)}$ along the third mode and obtain $\mathcal{X}$ which has dimensionality  $1024\!\times\!7\!\times\!2$. As FV contains the first- and second-order statistics, we use a separate stream per each type of statistics, and a single stream for BoW. For the practical choice of BoW/FV pipelines, we use either a Fully Connected (FC) unit shown in Figure \ref{fig:stra} or a Convolutional (Conv) pipeline in Figure \ref{fig:strb}. Thus, we investigate the following hallucinating stream combinations: 
% (i) BoW-FC and FV-FC, or (ii) BoW-Conv and FV-Conv. 
(i) BoW-FC and FV-FC, (ii) BoW-Conv and FV-FC, or (iii) BoW-Conv and FV-Conv.
Where indicated, we also equip each stream with Power Normalization (PN). For specific PN realizations, we investigate AsinhE, SigmE, and AxMin variants from Remarks \ref{re:asinhe}, \ref{re:sigme} and \ref{re:axmin}. %Moreover, as alluded to in Figure \ref{fig:pipeline}, due to large size of ground-truth training FV representations, we use the count sketching detailed in Section \ref{sec:sketch} at the training stage to match the size of each statistic to be 1000. Where indicated, we use count sketching for BoW although our ground-truth training BoW repr. are mostly 1000 dim. %We describe the details of the Means Square Loss applied to each stream in Section \ref{sec:obj}. 
Below we detail how we obtained  ground-truth BoW/FV. % descriptors.

\vspace{0.05cm}
\noindent{\textbf{Ground-truth BoW/FV.}} To train Fisher Vectors, we computed 256 dimensional GMM-based dictionaries on descriptors resulting from IDT \cite{improved_traj} according to steps described in Sections  \ref{sec:related} and \ref{sec:enc}. We applied PCA to trajectories (30 dim.), HOG (96 dim.), HOF (108 dim.), MBHx (96 dim.) and MBHy (96 dim.), and we obtained the final 213 dim. local descriptors. 
We applied encoding as in Eq. \eqref{eq:fisher1} and \eqref{eq:fisher2}, the aggregation from Section \ref{sec:aggr} and Power Normalization from Section \ref{sec:pns}. Thus, our encoded first- and second-order FV representations, each of size $256\!\times\!213\!=\!54528$, had to be sketched to 1000 dimensions. To this end, we followed Section \ref{sec:sketch}, prepared matrices $\mPP_{(fv1)}$ and $\mPP_{(fv2)}$ as in Proposition \ref{pr:ten_sketch}, and fixed both of them throughout experiments. The sketched first- and second-order representations  $\vec{\psi}'_{(fv1)}\!=\!\mPP_{(fv1)}\vec{\psi}_{(fv1)}$ and $\vec{\psi}'_{(fv2)}\!=\!\mPP_{(fv2)}\vec{\psi}_{(fv2)}$ can be readily combined next with the MSE loss functions detailed  in Section \ref{sec:obj}.

For BoW, we followed  Section \ref{sec:enc} and  applied k-means to build a 1000 dim. dictionary from the same descriptors which were employed to pre-compute FV. Then, the descriptors were encoded according to Eq. \eqref{eq:sp1}, aggregated according to steps described in Section \ref{sec:aggr} and normalized by Power Normalization from Section \ref{sec:pns}. Where indicated, we used 4000 dim. dictionary and thus applied sketching on such BoW to limit its vector size to 1000 dim. %To this end, we followed steps analogous to FV sketching procedure.

We note that we use ground-truth BoW/FV descriptors only at the training stage to train our hallucination streams. % At the testing time, where indicated, we use ground-truth BoW/FV only for comparisons, \etc.

\subsection{High Abstraction Features}
\label{sec:haf}

High Abstraction Features (HAF) take as input the I3D intermediate representations $\mathcal{X}_{(rgb)}$ and $\mathcal{X}_{(opt.)}$. Practical realizations of HAF pipelines are identical to those of BoW/FV/OFF. Thus, we have a choice of either FC or Conv units illustrated in Figures \ref{fig:stra} and \ref{fig:strb}. We simply refer to those variants as HAF-FC and HAF-Conv, respectively. Similar to BoW/FV/OFF streams, the HAF representation also uses Power Normalization and it is of size 1000.

\subsection{Optical Flow Features}
\label{sec:off}
For pipeline in Figure \ref{fig:pipeline2}, the I3D intermediate representation $\mathcal{X}_{(rgb)}$ only is fed to hallucination/HAF streams. I3D Optical Flow Features $\mathcal{X}_{(opt.)}$ are pre-computed as the training ground-truth for the OFF layer (the MSE loss is used).

\subsection{Combining Hallucinated BoW/FV/OFF and HAF$\!\!\!\!\!$}
\label{sec:concat}

Figure \ref{fig:pipeline} indicates that FV (first- and second-order), BoW and HAF feature vectors $\vec{\tilde{\psi}}_{(fv1)}$, $\vec{\tilde{\psi}}_{(fv2)}$, $\vec{\tilde{\psi}}_{(bow)}$ and $\vec{\psi}_{(haf)}$ are concatenated (via operator $\oplus$) to obtain $\vec{\psi}_{(tot)}$ and subsequently sketched  (if indicated so during experiments), that is, $\vec{\psi}'_{(tot)}\!=\!\mPP_{(tot)}\vec{\psi}_{(tot)}$ which reduces the size of the total representation from $d\!=\!4000$ to $500\!\leq\!d'\!\leq\!2000$. Matrix $\mPP_{(tot)}$ is prepared according to Proposition \ref{pr:ten_sketch} and fixed throughout experiments. As sketching is a linear projection, we can backpropagate through it with ease. When also hallucinating OFF as in Figure \ref{fig:pipeline2}, we additionally concatenate $\vec{\psi}_{(off)}$ with other feature vectors to obtain $\vec{\psi}_{(tot)}$.

\vspace{0.05cm}
\noindent{\textbf{PredNet.}} The final unit of our overall pipeline, PredNet, is illustrated in Figure \ref{fig:strc}. On input, we take $\vec{\psi}_{(tot)}$ (no sketching) or ($\vec{\psi}'_{(tot)}$) if sketching is used, pass it via the batch normalization and then an FC layer which produces a $C$ dim. representation passed to the cross-entropy loss.

\subsection{Objective and its Optimization}
\label{sec:obj}

%To train our pipeline
During training, we combine MSE loss functions responsible for training hallucination streams  with the class. loss:

\vspace{-0.4cm}
\begin{align}
&\;\;\!\!\!\!\!\!\!\!\!\!\!\ell^*(\tX, \vy; \bar{\mP})\!=\!\frac{\alpha}{|\mathcal{H}|}\sum\limits_{i\in\mathcal{H}}{\big\lVert{ \vec{\tilde{\psi}}_i\!-\!\vpsi'_i\big\rVert}}_2^2 \!+\! \ell\!\left(f\!(\vpsi'_{(tot)}; \mP_{(pr)}),\vy; \mP_{(\ell)}\right), \nonumber\\
&\qquad\qquad\text{ where: } \forall i\!\in\!\mathcal{H}, \vec{\tilde{\psi}}_i\!=\!g\!\left(\hslash(\tX, \mP_{i}),\eta\right),%\nonumber\\
%&\qquad\qquad\qquad\qquad\;\,
\vpsi'_i\!=\!\mPP_i\vpsi_i,\nonumber\\
&\qquad\qquad\qquad\quad\;\vpsi_{(haf)}\!=\!g\!\left(\hslash(\tX,\mP_{(haf)}),\eta\right),\nonumber\\
&\qquad\qquad\qquad\quad\;\vpsi'_{(tot)}\!=\!\mPP_{(tot)}\left[\oplus_{i\in\mathcal{H}}\vec{\tilde{\psi}}_i; \vpsi_{(haf)}\right].%\nonumber\\
%&\qquad\qquad\quad\;\mathcal{H}\equiv\left\{(fv1),(fv2),(bow)\right\}
\label{eq:loss}
\end{align}
The above equation is a trade-off between the MSE loss functions $\{{\lVert{ \vec{\tilde{\psi}}_i\!-\!\vpsi'_i\rVert}}_2^2, i\!\in\!\mathcal{H}\}$ and the classification loss $\ell(\cdot,\vy; \mP_{(\ell)})$ with some label $\vy\!\in\mathcal{Y}$ and parameters $\mP_{(\ell)}\!\equiv\!\{\mW,\vb\}$. The trade-off is controlled by a constant $\alpha\!\geq\!0$ while MSE is computed over hallucination streams $i\!\in\!\mathcal{H}$, and $\mathcal{H}\!\equiv\!\left\{(fv1),(fv2),(bow),(off)\right\}$ is our set of hallucination streams which can be modified to multiple/few such streams depending on the task at hand. Moreover, $g(\cdot,\eta)$ is a Power Normalizing function chosen from the family described in Section \ref{sec:pns}, $f(\cdot; \mP_{(pr)})$ is the PredNet module with parameters $\mP_{(pr)}$ which we learn, $\{\hslash(\cdot, \mP_{i}), i\!\in\!\mathcal{H}\}$ are the hallucination streams while $\{\vec{\tilde{\psi}}_i, i\!\in\!\mathcal{H}\}$ are the corresponding hallucinated BoW/FV/OFF representations. Moreover, $\hslash(\cdot, \mP_{(haf)})$ is the HAF stream with the output denoted by $\vpsi_{(haf)}$. For the hallucination streams, we learn parameters $\{\mP_{i}, i\!\in\!\mathcal{H}\}$ while for HAF, we learn  $\mP_{{(haf)}}$. The full set of parameters we learn is defined as $\bar{\mP}\!\equiv\!(\{{\mP_{i}, i\!\in\!\mathcal{H}}\}, \mP_{(haf)}, \mP_{(pr)},\mP_{(\ell)})$. Furthermore, $\{\mPP_{i}, i\!\in\!\mathcal{H}\}$  are the projection matrices for count sketching of the ground-truth BoW/FV/OFF feature vectors $\{\vpsi_{i}, i\!\in\!\mathcal{H}\}$ while $\{\vpsi'_i, i\!\in\!\mathcal{H}\}$ are the corresponding sketched/compressed representations. Finally, $\mPP_{(tot)}$ is the projection matrix for hallucinated BoW/FV/OFF representations concatenated with each other and HAF, that is, for $\vpsi_{(tot)}\!=\!\left[\oplus_{i\in\mathcal{H}}\vec{\tilde{\psi}}_i; \vpsi_{(haf)}\right]$ which results in the sketched counterpart $\vpsi'_{(tot)}$ that goes into the PredNet module $f$. Section \ref{sec:sketch} details how to select matrices $\mPP$. 
If sketching is not needed, we simply set a given $\mPP$ to be the identity projection $\mPP\!=\!\mIdent$. In our experiments, we simply set $\alpha\!=\!1$.

\begin{figure}[t]%htbp % left bottom right top
\centering%%%%
\vspace{-0.3cm}
\centering\includegraphics[trim=0 0 0 0, clip=true,width=0.45\textwidth]{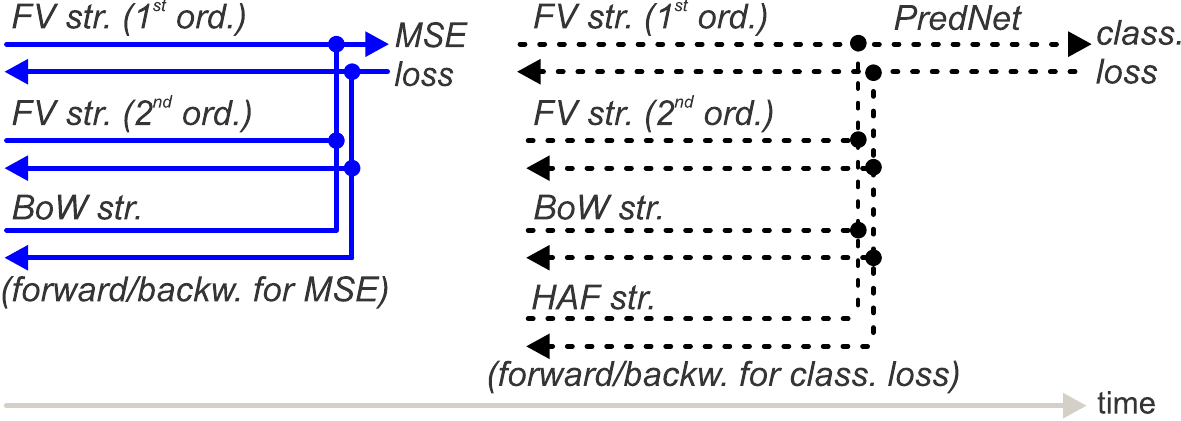}
\caption{Optimization. In each step, we have (i) forward/back\-ward passes via BoW/FV (optionally OFF) streams for the MSE loss followed by (ii) forward/backward passes via BoW/FV (opt. OFF), and HAF streams and PredNet for the classification loss.
}
\vspace{-0.3cm}
\label{fig:passes}
\end{figure}

\vspace{0.05cm}
\noindent{\textbf{Optimization.}} We minimize $\ell^*(\tX, \vy; \bar{\mP})$ w.r.t. parameters of each stream, that is $\{\mP_{i}, i\!\in\!\mathcal{H}\}$ for hallucination streams, $\mP_{(haf)}$ for the HAF stream, $\mP_{(pr)}$ for PredNet and $\mP_{(\ell)}$ for the classification loss. In practice, we perform a simple alternation over two minimization steps shown in Figure \ref{fig:passes}. In each iteration, we perform one forward and backward pass regarding the MSE losses to update the parameters  $\{\mP_{i}, i\!\in\!\mathcal{H}\}$ of the hallucination streams. Then, we perform one forward and backward pass regarding the classification loss $\ell$. We update all network streams during this pass. Thus, one can think of our network as multi-task learning  with BoW/FV/OFF and label learning tasks. Furthermore, we use the Adam minimizer with $10^{-4}$ initial learning rate which we halve every 10 epochs. We run our training for 50--100 epochs depending on the dataset.

\vspace{0.05cm}
\noindent{\textbf{Sketching the Power Normalized vectors.}}
\begin{proposition}
\vspace{-0.2cm}
\label{prop:fact}
Sketching PN vectors increases the sketching variance ($\ell_2$-normalized by vec. norms)  by  $1\!\leq\!\kappa\!\leq\!2$.
\begin{proof}
\vspace{-0.3cm}
Normalize variance $\mathbb{V}$ from Remark \ref{re:ten_sketch} by the norms $\lVert\vpsi\rVert_2^2\lVert\vpsi'\rVert_2^2$. Consider $\mathbb{V}^{(\gamma)}$ which is the variance for $d$ dimensional vectors $\{(\vpsi^\gamma,\vpsi'^\gamma)\!:\!\vpsi\!\geq\!0,\vpsi'\!\geq\!0\}$ power normalized by Gamma from Remark \ref{re:asinhe}, and divide it accordingly by $\lVert\vpsi^\gamma\rVert_2^2\lVert\vpsi'^\gamma\rVert_2^2$. For extreme PN ($\gamma\!\rightarrow\!0$), we have:
\vspace{-0.2cm}
\begin{equation}
\!\!\!\!\!\!\!\!\!\!\!\!\lim\limits_{\gamma\!\rightarrow\!0}\mathbb{V}^{(\gamma)}\!=\!\frac{1}{d'}\lim\limits_{\gamma\!\rightarrow\!0}\left(\frac{\left<\vpsi^\gamma,\vpsi'^\gamma\right>^2}{\lVert\vpsi^\gamma\rVert_2^2\lVert\vpsi'^\gamma\rVert_2^2}\!+\!1\right)\!=%\!\frac{1}{d'}(\frac{d^2}{d^2}\!+\!1)\!=
\!\frac{2}{d'}.
\end{equation}
Now, assume that $d$ dimensional $\vpsi$ and $\vpsi'$ are actually $\ell_2$-norm normalized. Then, we have the following ratio of variances:
\vspace{-0.2cm}
\begin{equation}
%\kappa\!=\!\mathbb{V}/\mathbb{V}^{(\gamma)}=\frac{2}{\left<\vpsi,\vpsi'\right>^2\!+\!1},
\kappa\!=\!\mathbb{V}/\mathbb{V}^{(\gamma)}=2/({\left<\vpsi,\vpsi'\right>^2\!+\!1}),
\end{equation}
Thus, $1\!\leq\!\kappa\!\leq\!2$ depends on $(\vpsi,\vpsi')$, and $\kappa$ varies smoothly between $[1; 2]$ for $1\!\leq\!\gamma\!\leq\!0$ of Gamma, a monotonically increasing function. For typical $\gamma\!=\!0.5$, we measured for the actual data that $\kappa\!\approx\!1.3$.
\end{proof}
\end{proposition}
\comment{
\begin{figure}[t]%htbp % left bottom right top
%\hspace{0.5cm}%
\centering%%%%
\vspace{-0.3cm}
\begin{subfigure}[b]{0.45\linewidth}
\centering\includegraphics[trim=0 0 0 0, clip=true,width=0.95\linewidth]{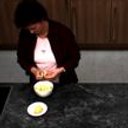}
\vspace{-0.2cm}
%\caption{\label{fig:stra}}
%\vspace{0.2cm}
\end{subfigure}
\begin{subfigure}[b]{0.45\linewidth}
\centering\includegraphics[trim=0 0 0 0, clip=true,width=0.95\linewidth]{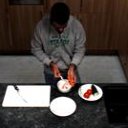}
\vspace{-0.2cm}
%\caption{\label{fig:stra}}
%\vspace{0.2cm}
\end{subfigure}
\caption{MPII Cooking Activities. Note the fine-grained  tasks.
}\vspace{-0.3cm}
\label{fig:cook}
\end{figure}
}

\section{Experiments}
\label{sec:exper}

%Below we describe datasets/protocols, and  evaluations. %We do not stake claims that we beat every single AR pipeline nor this is our goal. %However, we make our work complementary to 
%As many AR pipelines use IDT descriptors to boost their performance \cite{cnn3d_ar,i3d_net,hok,anoop_rankpool_nonlin,anoop_advers}, in this paper, we simply show how to put new life into handcrafted IDT descriptors. We learn and synthesize them (and optical flow features) by a CNN pipeline. %Firstly, we describe datasets and evaluation protocols.

\subsection{Datasets and Evaluation Protocols}
\label{sec:data}

\noindent\textbf{HMDB-51~\cite{kuehne2011hmdb}} %is a popular video benchmark for human action recognition, 
consists of 6766 internet videos over 51 classes; each video has $\sim$20--1000 frames. %The standard evaluation protocol reports average classification accuracy on three-fold splits. 
Following the protocol, we report the mean accuracy across three splits. %Where stated, we often investigate various hyperparameters on the first split.

\noindent\textbf{YUP++~\cite{yuppp}} dataset contains so-called  video textures. It has 20 scene classes, 60 videos per class, and its splits contain scenes captured with the static or moving camera. We follow the standard splits (1/9 dataset for training). %We report result for the static, dynamic and mixed capture sequences.

\noindent\textbf{MPII Cooking Activities~\cite{rohrbach2012database}} consist of high-resolution videos of people cooking various dishes. %; each video contains a single person cooking a dish. %and overall there are 12 such videos in the dataset. 
The 64 distinct activities from 3748 clips include coarse actions \eg, \emph{opening refrigerator}, and fine-grained actions \eg, \emph{peel}, \emph{slice}, \emph{cut apart}.
 %and one background activity (1861 clips).
% (see Figure \ref{fig:cook}). %This dataset is challenging due to (i) unbalanced action classes, (ii) significant intra-class differences (each subject cooks according to their own style).  
We use the mean Average Precision (mAP) over 7-fold cross validation. For human-centric protocol \cite{anoop_generalized,anoop_rankpool_nonlin}, we use Faster-RCNN \cite{faster-rcnn} to crop video around humans.

\noindent\textbf{Charades~\cite{sigurdsson2016hollywood}} consist of of 9848 videos of daily indoors activities, 66500 temporal annotations and 157 classes.

\begin{table}[t]%htbp % left bottom right top
%\renewcommand{\arraystretch}{0.65}
%\fontsize{8}{8}\selectfont
%\footnotesize
\vspace{-0.3cm}
\parbox{.99\linewidth}{
\setlength{\tabcolsep}{0.12em}
\renewcommand{\arraystretch}{0.70}
%\fontsize{9}{9}\selectfont
\centering
\begin{tabular}{ c | c | c | c | c }
 & {\em sp1} & {\em sp2} & {\em sp3} & mean acc. \\
\hline
HAF only     				& $81.83\%$ & $80.78\%$ & $80.45\%$ & $81.02\%$\\
HAF+BoW/FV exact    & $83.00\%$ & $82.80\%$ & $81.70\%$ & $\mathbf{82.50}\%$\\
\hline
HAF+BoW halluc.     & $82.29\%$ & $81.24\%$ & $80.98\%$ & $81.50\%$\\
HAF+FV halluc.     	& $82.68\%$ & $81.05\%$ & $79.93\%$ & $81.22\%$\\
HAF+BoW/FV halluc.  & $82.88\%$ & $82.74\%$ & $81.50\%$ & $\mathbf{82.37}\%$\\
\hline
\end{tabular}
%\vspace{0.2cm}\\
}
\caption{Evaluations of pipelines on the HMDB-51 dataset. We compare ({\em HAF only}) and ({\em HAF+BoW/FV exact}) which show the lower- and upper bound on the accuracy, and our ({\em HAF+BoW/FV halluc.}), ({\em HAF+BoW halluc.}) and ({\em HAF+FV halluc.}).}
\vspace{-0.2cm}
\label{tab:hmdb51a}
\end{table}

\begin{table}[t]%htbp % left bottom right top
%\renewcommand{\arraystretch}{0.65}
%\fontsize{8}{8}\selectfont
%\footnotesize
\parbox{.99\linewidth}{
\setlength{\tabcolsep}{0.12em}
\renewcommand{\arraystretch}{0.70}
%\fontsize{9}{9}\selectfont
\centering
\begin{tabular}{ c | c | c | c | c }
 & {\em static} & {\em dynamic} & {\em mixed} & mean acc. \\
\hline
HAF only     				& $92.03\%$ & $81.67\%$ & $89.07\%$ & $87.59\%$\\
HAF+BoW/FV exact    & $93.30\%$ & $89.82\%$ & $92.41\%$ & $91.84\%$\\
\hline
HAF+BoW halluc.     & $92.69\%$ & $85.93\%$ & $92.41\%$ & $90.34\%$\\
HAF+FV halluc.     	& $92.69\%$ & $88.15\%$ & $91.48\%$ & $90.77\%$\\
HAF+BoW/FV halluc.  & $93.15\%$ & $89.63\%$ & $92.31\%$ & $\mathbf{91.69}\%$\\
\hline
\end{tabular}
%\vspace{0.2cm}\\
}
\caption{Eval. of pipelines on  YUP++. %We compare ({\em HAF only}) and ({\em HAF+BoW/FV exact}) which show the lower- and upper bound on the accuracy, and our ({\em HAF+BoW/FV halluc.}), ({\em HAF+BoW halluc.}) and ({\em HAF+FV halluc.}).
See Table \ref{tab:hmdb51a} for the legend.}
\vspace{-0.2cm}
\label{tab:yupa}
\end{table}

\begin{table}[t]%htbp % left bottom right top
%\renewcommand{\arraystretch}{0.65}
%\fontsize{8}{8}\selectfont
%\footnotesize
%\vspace{-0.3cm}
\hspace{-0.3cm}
\parbox{.99\linewidth}{
\setlength{\tabcolsep}{0.12em}
\renewcommand{\arraystretch}{0.70}
\fontsize{9}{9}\selectfont
\centering
\begin{tabular}{ l | c | c | c | c }
 & {\em sp1} & {\em sp2} & {\em sp3} & mean acc. \\
\hline
HAF-Conv+BoW/FV-FC halluc.  & $81.96\%$ & $80.39\%$ & $80.52\%$ & $80.95\%$\\
HAF-FC+BoW/FV-Conv halluc.  & $82.42\%$ & $81.30\%$ & $81.50\%$ & $81.74\%$\\
HAF-FC+BoW/FV-FC halluc.  & $82.88\%$ & $82.74\%$ & $81.50\%$ & $\mathbf{82.37}\%$\\
\hline
\end{tabular}
%\vspace{0.2cm}
}
\caption{Evaluations of pipelines on the HMDB-51 dataset. We compare ({\em HAF+BoW/FV halluc.}) approach on different architectures used for HAF and BoW/FV streams such as ({\em FC}) and ({\em Conv}).}
\vspace{-0.3cm}
\label{tab:hmdb51b}
\end{table}

\subsection{Evaluations}
\label{sec:evals}

We start our experiments by investigating various aspects of our pipeline and then we present our final results.

\vspace{0.05cm}
\noindent{\textbf{HAF, BoW and FV streams. }} Firstly, we ascertain the gains from our HAF and BoW/FV streams. We evaluate the performance of (i) the HAF-only baseline pipeline without IDT-based BoW/FV information ({\em HAF only}), (ii) the HAF-only baseline with exact ground-truth  IDT-based BoW/FV added at both training and testing time ({\em HAF+BoW/FV exact}), and (iii) the combined HAF plus IDT-based BoW/FV streams ({\em HAF+BoW/FV halluc.}). We also perform evaluations on (iv) HAF  plus IDT-based BoW stream ({\em HAF+BoW halluc.}) and HAF plus IDT-based FV stream ({\em HAF+FV halluc.}) to examine how much gain IDT-based BoW and FV bring, respectively. As Section \ref{sec:fv-bow} suggests that each stream can be based on either the Fully Connected (FC) or Convolutional (Conv.) pipeline, we firstly investigate the use of FC unit from Figure \ref{fig:stra}, that is, we use HAF-FC, BoW-FC and HAF-FC streams. PredNet also uses FC. For ground-truth FV, we use 1000 dim. sketches. %per first- and second-order components.

Table \ref{tab:hmdb51a} presents results on the HMDB-51 dataset. As expected, the ({\em HAF only}) is the poorest performer while ({\em HAF+BoW/FV exact}) is the best performer determining the upper limit on the accuracy. Hallucinating  ({\em HAF+BoW halluc.}) outperforms ({\em HAF+FV halluc.}) marginally. We expect FV to perform close to BoW  due to the significant compression with sketching by factor $\sim\!52.5\!\times$. Approaches ({\em HAF+FV/BoW halluc.}) and  ({\em HAF+BoW/FV exact}) achieve the best results, and outperform ({\em HAF only}) by 1.35\% and  1.48\% accuracy. These result show that our hallucination strategy ({\em HAF+FV/BoW halluc.}) can mimic  ({\em HAF+BoW/FV exact}) closely. Our $82.37$\% accuracy is the new state of the art. Below we show larger gains on YUP++.

Table \ref{tab:yupa} presents similar findings on the YUP++ dataset. Our ({\em HAF+FV halluc.}) brings the improvement of $\sim\!2.2$ and $\sim\!6.5$\% over ({\em HAF+BoW halluc.}) and ({\em HAF only}) on scenes captured with the moving camera ({\em dynamic}). Our ({\em HAF+BoW/FV halluc.}) yields $\sim\!8.0$\% over ({\em HAF only}) thus demonstrating again the benefit of hallucinating BoW/FV descriptors. The total gain for ({\em HAF+BoW/FV halluc.}) over ({\em HAF only}) equals $4.1$\%. Our  ({\em HAF+FV/BoW halluc.}) matches results of  ({\em HAF+BoW/FV exact}) without explicitly computing BoW/FV during testing. Below, we  investigate different architectures of our streams.

\vspace{0.05cm}
\noindent{\textbf{Fully Connected/Convolutional streams. }} Figures \ref{fig:stra} and \ref{fig:strb} show two possible realizations of HAF, BoW and FV streams. While FC and Conv. architectures are not the only possibilities, they are the simplest ones. Table \ref{tab:hmdb51b} shows that using FC layers ({\em FC}) for HAF and BoW/FV streams, denoted as ({\em HAF-FC+BoW/FV-FC halluc.}) outperforms Convolutional ({\em Conv}) variants by up to $\sim\!1.5$\% accuracy. Thus, we use only the FC architecture in what follows.

\vspace{0.05cm}
\noindent{\textbf{Sketching and Power Normalization. }}%Below, we investigate the impact of Power Normalization and count sketching presented in Sections  \ref{sec:pns} and \ref{sec:sketch} on the accuracy. 
As PredNet uses FC layers (see Figure \ref{fig:strc}), we expect that limiting the input size to this layer via count sketching from Section \ref{sec:sketch} should benefit the performance. Moreover, as visual and video representations suffer from so-called burstiness, we  investigate AsinhE, SigmE and AxMin from Remarks \ref{re:asinhe}, \ref{re:sigme} and \ref{re:axmin}.

\begin{figure}[b]%htbp % left bottom right top
\vspace{-0.5cm}
\hspace{-0.3cm}
\centering%%%%
\begin{subfigure}[b]{0.49\linewidth}
\centering\includegraphics[trim=0 0 0 0, clip=true,width=1.1\linewidth]{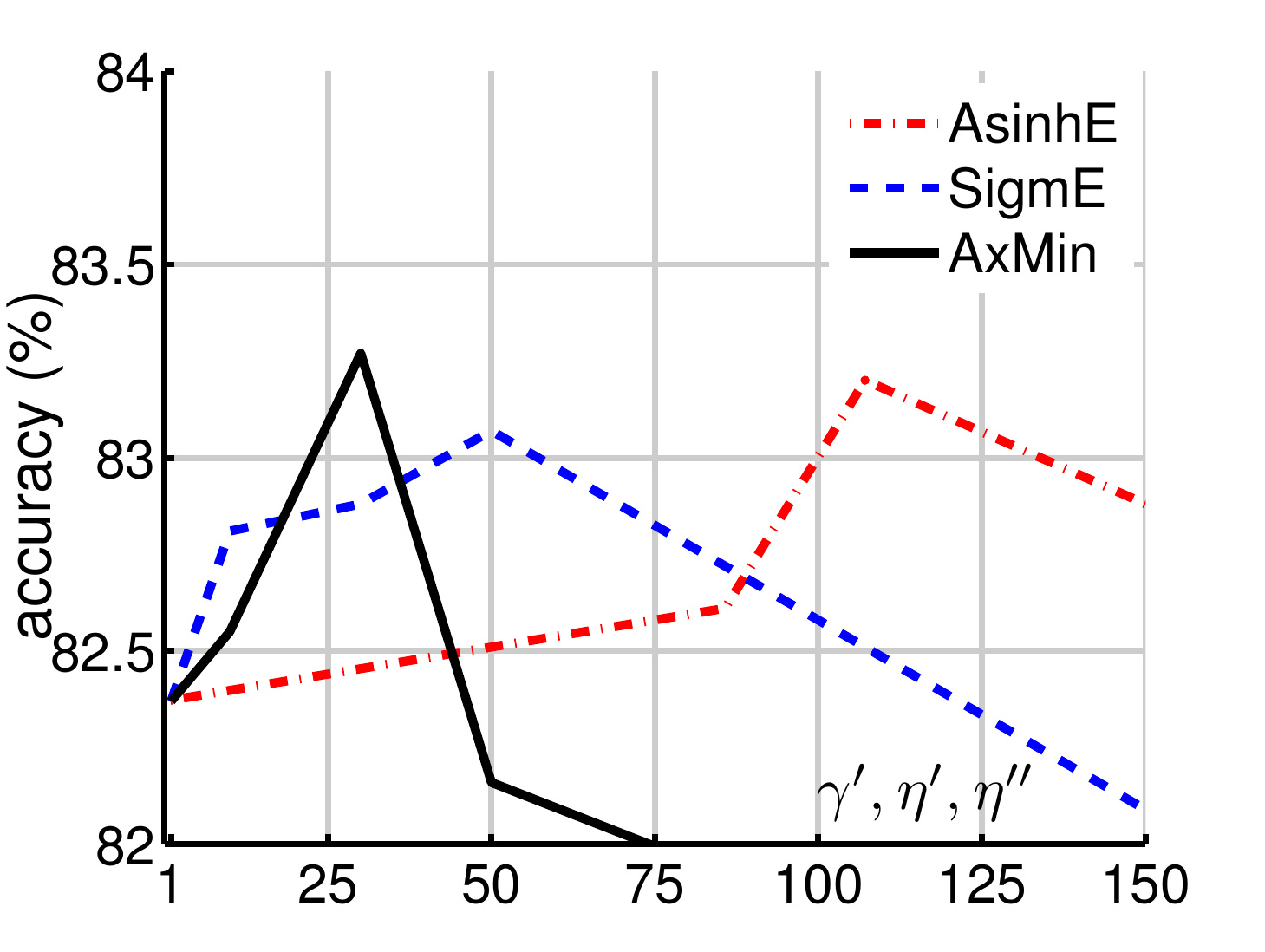}
\vspace{-0.6cm}
\caption{\label{fig:pn}}
\vspace{-0.2cm}
\end{subfigure}
\hspace{0.1cm}
\begin{subfigure}[b]{0.49\linewidth}
\centering\includegraphics[trim=0 0 0 0, clip=true,width=1.1\linewidth]{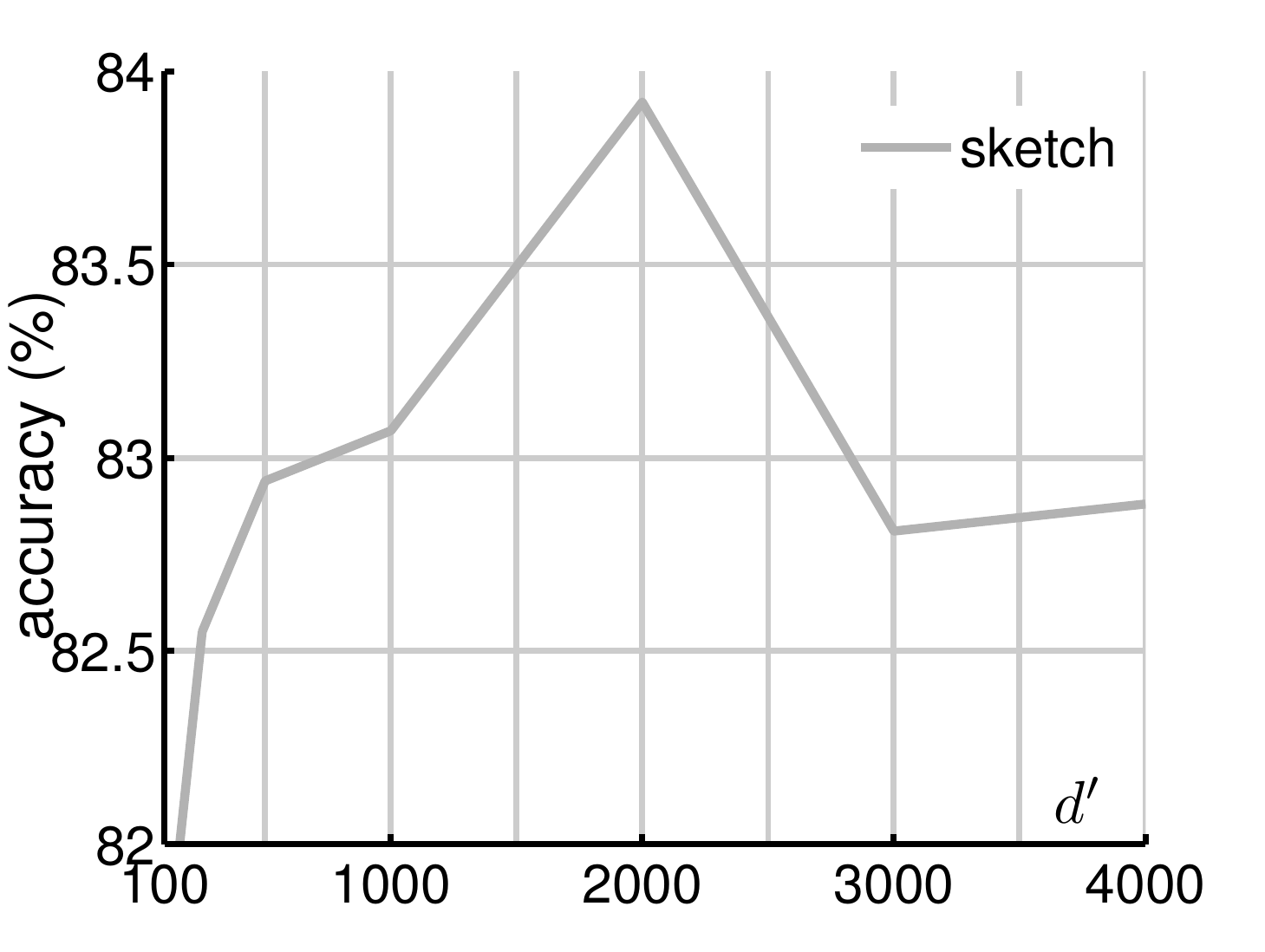}
\vspace{-0.6cm}
\caption{\label{fig:sketch}}
\vspace{-0.2cm}
\end{subfigure}
%
%\vspace{-0.7cm}
\caption{Evaluations of (fig. \ref{fig:pn}) Power Normalization and (fig. \ref{fig:sketch})  sketching on the HMDB-51 dataset (split 1 only). 
}\vspace{-0.3cm}
\label{fig:pnsk}
\end{figure}

Figure \ref{fig:pn} investigates the classification accuracy on the HMDB-51 dataset (split 1) when our HAF and BoW/FV feature vectors $\{\vec{\tilde{\psi}}_i, i\!\in\!\mathcal{H}\}$ and $\vpsi_{(haf)}$ (described in Sections \ref{sec:concat} and \ref{sec:obj}) are passed via Power Normalizing functions  AsinhE, SigmE and AxMin prior to concatenation (see Figure \ref{fig:pipeline}). From our experiment it appears that all PN functions perform similarly and improve results from the baseline $82.29$\% to $\sim\!83.20$\% accuracy. We observe a similar gain from $93.15$\% to $94.44$\% acc. on  YUP++ ({\em static}). In what follows, we simply use AsinhE for PN.

Figure \ref{fig:sketch} illustrates on the HMDB-51 dataset (split 1) that applying count sketching on concatenated HAF and BoW/FV feature vectors $\vpsi_{(tot)}$, which produces $\vpsi'_{(tot)}$ (see Section \ref{sec:obj} for reference to symbols), improves results from $82.88$\% to $83.92$\% accuracy for $d'\!=\!2000$. This is expected as reduced size of $\vpsi'_{(tot)}$ results in fewer parameters of the FC layer of PredNet and less overfitting. Similarly,  for the YUP++ dataset and the split ({\em static}), we see the performance increase from $93.15$\% to $94.81$\% accuracy.

\vspace{0.05cm}
\noindent{\textbf{Comparisons with other methods. }} Below we present our final results and we contrast them against the state of the art. Table \ref{tab:hmdb51f} shows results on the HMDB-51 dataset. For our method, we used sketching of $\vpsi_{(tot)}$ with $d'\!=\!2000$ and PN. Our ({\em HAF+BoW/FV halluc.}) model yields $82.48$\% acc. which beats  results in the literature to the best of our knowledge. If we tune PN per split, our results reach  $82.78$\% accuracy. However, we do not advise such tuning due to danger of overfitting. We note that we outperform more complex methods such as Adversarial Discriminative Learning (ADL) with I3D \cite{anoop_advers} and Fully Fine-Tuned I3D \cite{i3d_net}.

\begin{table}[t]%htbp % left bottom right top
\vspace{-0.3cm}
\parbox{.99\linewidth}{
\setlength{\tabcolsep}{0.12em}
\renewcommand{\arraystretch}{0.70}
%\fontsize{9}{9}\selectfont
\centering
\begin{tabular}{ c | c | c | c | c }
 & {\em sp1} & {\em sp2} & {\em sp3} & mean acc. \\
\hline
HAF only     				& $81.83\%$ & $80.78\%$ & $80.45\%$ & $81.02\%$\\
HAF+BoW/FV halluc.  & $83.46\%$ & $82.61\%$ & $81.37\%$ & $\mathbf{82.48}\%$\\
\hline
\end{tabular}
\vspace{0.02cm}\\
}
\parbox{.99\linewidth}{
\setlength{\tabcolsep}{0.12em}
\renewcommand{\arraystretch}{0.70}
\fontsize{9}{9}\selectfont
\centering
\begin{tabular}{ c | c }
\hline
\kern-0.5em ADL+ResNet+IDT $74.3\%$ \cite{anoop_advers} & STM Network+IDT $72.2\%$ \cite{stm_net}\kern-0.5em\\
\kern-0.5em ADL+I3D $81.5\%$ \cite{anoop_advers} &  Full-FT I3D $81.3\%$ \cite{i3d_net}\\
\hline
\end{tabular}
}
\caption{Evaluations of ({\em top}) our ({\em HAF+BoW/FV halluc.}) and ({\em bottom}) comparisons to the state of the art on  HMDB-51.}
\vspace{-0.2cm}
\label{tab:hmdb51f}
\end{table}

\begin{table}[t]%htbp % left bottom right top
%\vspace{-0.3cm}
\parbox{.99\linewidth}{
\setlength{\tabcolsep}{0.12em}
\renewcommand{\arraystretch}{0.70}
%\fontsize{9}{9}\selectfont
\centering
\begin{tabular}{ c | c | c | c || c | c}
 & \multirow{2}{*}{\em static} & \multirow{2}{*}{\em dynamic} & \multirow{2}{*}{\em mixed} & mean & mean \\
&                &             &             & {\fontsize{8}{9}\selectfont stat/dyn}  & all \\
\hline
HAF only     				& $92.03\%$ & $81.67\%$ & $89.07\%$ & $86.8\%$ & $87.6\%$\kern-0.5em \\
\kern-0.5em {\fontsize{8}{9}\selectfont HAF+BoW/FV halluc.}  & $94.81\%$ & $\mathbf{89.63}\%$ & $\mathbf{93.33}\%$ & $\mathbf{92.2}\%$ & $\mathbf{92.6}\%$\kern-0.5em \\
\hline
\hline
T-ResNet \cite{yuppp} & $92.41\%$ & $81.50$\% & $89.00$\% & $87.0\%$ & $87.6\%$\\
ADL I3D \cite{anoop_advers} & $\mathbf{95.10}\%$ & $88.30$\% & - & $91.7\%$ & -\\
\hline
\end{tabular}
%\vspace{0.2cm}\\
}
\caption{Evaluations of ({\em top}) our ({\em HAF+BoW/FV halluc.}) and ({\em bottom}) comparisons to the state of the art on YUP++.}
\vspace{-0.2cm}
\label{tab:yupf}
\end{table}

\begin{table}[!tb]%htbp % left bottom right top
\parbox{.99\linewidth}{
\setlength{\tabcolsep}{0.12em}
\renewcommand{\arraystretch}{0.70}
%\fontsize{9}{9}\selectfont
\hspace{-0.3cm}
%\centering
\begin{tabular}{ c | c | c | c | c | c | c | c | c }
 & {\em sp1} & {\em sp2} & {\em sp3} & {\em sp4} & {\em sp5} & {\em sp6} & {\em sp7} & mAP \\
\hline
\kern-0.5em {\fontsize{8}{9}\selectfont HAF+BoW halluc.}      		 & $73.9$ & $71.6$ & $76.2$ & $70.7$ & $76.3$ & $71.9$ & $63.4$ & $71.9\%$\\
\kern-0.5em {\fontsize{7.5}{9}\selectfont HAF+BoW halluc.+SK/PN}     & $73.9$ & $75.8$ & $72.2$ & $73.9$ & $77.0$ & $73.6$ & $68.8$ & $\mathbf{73.6}\%$\\
\hline
\kern-0.5em {\fontsize{8}{9}\selectfont HAF* only}      		     & $74.6$ & $73.2$ & $77.0$ & $75.1$ & $76.1$ & $75.6$ & $71.9$ & $74.8\%$\\
\kern-0.5em {\fontsize{8}{9}\selectfont HAF*+BoW halluc.}      		 & $78.8$ & $75.0$ & $84.1$ & $76.0$ & $77.0$ & $78.3$ & $75.2$ & $\mathbf{77.8}\%$\\
\kern-0.5em {\fontsize{7.5}{9}\selectfont HAF*+BoW hal.+MSK/PN}   & $80.1$ & $79.2$ & $84.8$ & $83.9$ & $80.9$ & $78.5$ & $75.5$ & $\mathbf{80.4}\%$\\
\hline
\kern-0.5em {\fontsize{7.5}{9}\selectfont HAF$^\bullet$+BoW hal.+MSK/PN}   & $80.8$ & $80.9$ & $85.0$ & $83.9$ & $82.0$ & $79.8$ & $79.6$ & $\mathbf{81.7}\%$\\
\kern-0.5em {\fontsize{7.5}{9}\selectfont ditto+OFF hal.}       		  & $81.2$ & $81.2$ & $84.9$ & $83.4$ & $84.2$ & $78.9$ & $79.1$ & $\mathbf{81.8}\%$\\
\hline
\kern-0.5em {\fontsize{7.5}{9}\selectfont I3D+BoW MTL$^\bullet$}       	& $79.1$ & $78.1$ & $83.6$ & $78.7$ & $79.1$ & $78.6$ & $76.5$ & $79.1$\%\\
\hline
\end{tabular}
\vspace{0.02cm}%\\
}
\parbox{.99\linewidth}{
\setlength{\tabcolsep}{0.12em}
\renewcommand{\arraystretch}{0.70}
\hspace{-0.45cm}
\fontsize{8}{9}\selectfont
%\centering
\begin{tabular}{ c | c | c | c}
\hline
%\kern-0.5em KRP-FS $70.0\%$ \cite{anoop_rankpool_nonlin} & KRP-FS+IDT $76.1\%$ \cite{anoop_rankpool_nonlin}\kern-0.5em\\
%\kern-0.5em GRP $68.4\%$ \cite{anoop_generalized} & GRP+IDT $75.5\%$ \cite{anoop_generalized}\kern-0.5em\\
\kern-0.5em KRP-FS $70.0\%$ \cite{anoop_rankpool_nonlin} & KRP-FS+IDT $76.1\%$ \cite{anoop_rankpool_nonlin} &
 GRP $68.4\%$ \cite{anoop_generalized} & GRP+IDT $75.5\%$ \cite{anoop_generalized}\kern-0.5em\\
\hline
\end{tabular}
}
\caption{Evaluations of ({\em top}) our ({\em HAF+BoW halluc.}) pipeline without sketching/PN, with sketching/PN ({\em SK/PN}). The ({\em HAF* only}) is our baseline without the BoW stream, ({\em *}) denotes human-centric pre-processing while ({\em MSK/PN})in pipeline ({\em HAF*+BoW hal.+MSK/PN}) denotes multiple sketches per BoW followed by Power Norm ({\em PN}). ({\em bottom}) Other methods on the MPII dataset.}
\vspace{-0.2cm}
\label{tab:mpiif}
\end{table}

\newcommand{\fsnine}[0]{\fontsize{9}{9}\selectfont}
\newcommand{\fsninee}[0]{\fontsize{7.5}{9}\selectfont}
\begin{table}[!tb]%htbp % left bottom right top
\fontsize{8}{8}\selectfont
%\footnotesize
\hspace{-0.525cm}
\parbox{.99\linewidth}{
\setlength{\tabcolsep}{0.12em}
\renewcommand{\arraystretch}{0.70}
%\fontsize{9}{9}\selectfont
\centering
\begin{tabular}{ c | c | c | c | c }
HAF     				& HAF+BoW/          & \fsninee HAF+BoW/FV/OFF   & \fsninee HAF+BoW/FV/OFF   & \fsninee HAF+BoW/FV/OFF   \\
only     				& FV exact          & halluc. +MSK$\times\!2$/PN & halluc. +MSK$\times\!4$/PN & halluc. +MSK$\times\!8$/PN \\
\hline
\fsnine 37.2 & \fsnine 41.9 & \fsnine 42.0 & \fsnine 42.2 & \fsnine\textbf{43.1} \\
\hline
\end{tabular}
%\vspace{0.2cm}\\
}
\caption{Evaluations of our methods on the Charades dataset.}
\vspace{-0.3cm}
\label{tab:charades}
\end{table}

Table \ref{tab:yupf} shows results on the YUP++ dataset. Our ({\em HAF+BoW/FV halluc.}) model yields very competitive results on the static protocol and outperforms competitors on the dynamic and mixed protocols. With 92.2\% mean accuracy over static and dynamic scores ({\em mean stat/dyn}), we outperform more complex ADL+I3D  \cite{anoop_advers} and T-ResNet \cite{yuppp}.

Table \ref{tab:mpiif} shows results for the MPII dataset for which we use HAF with/without the BoW (4000 dim.) hallucination stream (no FV stream). As MPII contains subsequences, we use integral pooling from Prop. \ref{pr:subseq}.  Our basic model ({\em HAF+BoW halluc.}) scores $\sim\!71.9$\% mAP. Applying sketching and PN ({\em HAF+BoW halluc.+SK/PN}) yields $73.6$\% mAP. Unlike GRP+IDT \cite{anoop_generalized} and KRP-FS+IDT \cite{anoop_rankpool_nonlin}, our first two experiments do not use any human- or motion-centric pre-processing. With human-centric crops, denoted with ({\em*}), our baseline without BoW ({\em HAF* only}) achieves $74.8$\% mAP. The model with BoW ({\em HAF+BoW halluc.}) scores $77.8$\% mAP. By utilizing 4 sketches for BoW and 4 BoW streams with Power Normalization ({\em HAF*+BoW hal.+MSK/PN}), we obtain $80.4$\% mAP. %This shows that despite a slight increase in sketching variance due to PN preceding sketching, as explained in Prop. \ref{prop:fact}, preventing overfitting/bursitess in AR is vital.

\vspace{0.05cm}
\noindent{\textbf{Hallucinating Optical Flow. }} For ({\em HAF$^\bullet$+BoW hal.+MSK/$\!$\\PN}) in Table \ref{tab:mpiif}, we increased the resolution of RGB frames $2\!\times$ to obtain larger human-centric crops and $2\!\times$ larger  optical flow res., which yielded $81.7$\% mAP. In the same setting, hallucinating optical flow feat. ({\em ditto+OFF hal.}) yielded $81.84$\% mAP, the new state of the art.

\vspace{0.05cm}
\noindent{\textbf{Charades. }} 
In Table \ref{tab:charades}, baselines ({\em HAF only}) and ({\em HAF+Bo\\-W/FV exact}) score $37.2$\% and $41.9$\% mAP. Moreover, our best pipeline ({\em HAF+BoW/FV/OFF halluc.+MSK$\times\!8$/PN}) that hallucinates IDT BoW/FV and I3D optical flow features (OFF) with 8 sketches per BoW/FV/OFF and PN yielded $43.1\%$ (a much more complex feature banks \cite{Wu_2019_CVPR} yield $43.4\%$). Finally, if 25\% of this dataset was dedicated to
testing, $\sim$55h of computations would be saved.

%\vspace{0.05cm}
\noindent{\textbf{Discussion.}} 
%\section{Discussion}
%\label{sec:disc}
There exist several reasons explaining why our pipeline works well \eg, sophisticated  IDT trajectory modeling is
%IDT perform camera motion estimation and capture motion information along motion trajectories (tracked along sequence) while pruning inconsistent matches. Such operations are
 unlikely to be captured by off-the-shelf CNNs unless a CNN is enforced to learn IDT. We perform ‘translation’ of the
I3D output into IDT-based BoW/FV descriptors thus enforcing I3D to implicitly learn IDT which co-regularizes I3D which resembles Domain Adaptation (DA) methods: a source network co-regularizes a target network \cite{me_domain,ar_domain,Koniusz2018Museum,samitha_minmax,mer_spd_manifolds,Soumava_ICCV19} by the alignment of feature statistic of both streams. Related to DA is Multi-task Learning (MTL) known for boosting generalization/preventing overfitting of CNNs due to task specific losses \cite{Caruana-ML}. MTL training on related tasks is known to boost individual task accuracies beyond a vanilla feature fusion \cite{thrun_nips}. Finally, our pipeline uses self-supervision \eg, IDT BoW/FV and OFF descriptors represent easy to obtain self-information about videos. 
We train our halluc./last I3D layers via task-specific losses (similar to MTL). However, our halluc. layers distill the domain specific cues
which are fed back into the network (PredNet) which boosts our results by further $\sim$2.7\% compared to vanilla ({\em I3D+BoW MTL$^\bullet$}) in Table \ref{tab:mpiif}.
%
%
%However, we create a dedicated layer which hallucinates BoW/FV representations via an additional layer on top of I3D stream to use them at the test time thus further boosting results. 
%See Appendix \ref{sec:mse} for the quality of hallucination.

\section{Conclusions}
\label{sec:concl}

We have proposed a simple yet powerful strategy that learns IDT-based descriptors (and even optical flow features) and hallucinates them in a CNN pipeline for AR at the test time. With state-of-the-art results, we hope our method will spark a renewed interest in IDT-like descriptors. %Kindly see our supplementary material for further analysis/results.

\vspace{0.05cm}
\noindent{\textbf{Acknowledgements.}}  We thank CSIRO Scientific Computing for their help and NVIDIA for GPUs. We thank Dr. M. R. Mansour for  early discussion on related topics.

\begin{appendices}

%\begin{minipage}[c]{\textwidth}
\begin{figure*}[!htbp]%htbp % left bottom right top
%\hspace{0.5cm}%
\centering%%%%
\vspace{-0.3cm}
\comment{
\begin{subfigure}[b]{0.245\linewidth}
\centering\includegraphics[trim=0 0 0 0, clip=true,width=0.95\linewidth]{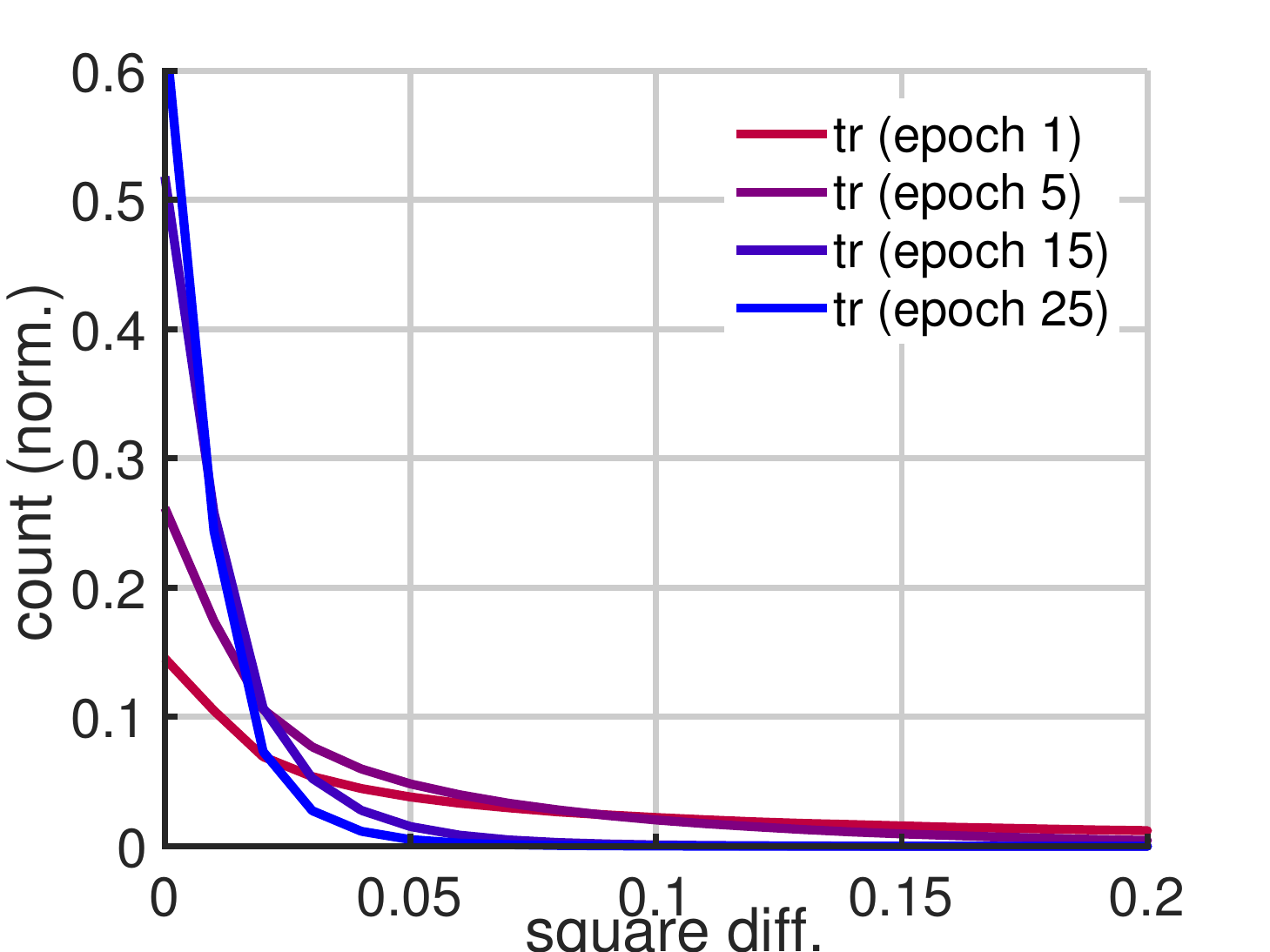}
%\vspace{-0.2cm}
\caption{\label{fig:str11} BoW FC (train)}
%\captionof{subfigure}{aaa\label{fig:stra}}
%\vspace{0.2cm}
\end{subfigure}
\begin{subfigure}[b]{0.245\linewidth}
\centering\includegraphics[trim=0 0 0 0, clip=true,width=0.95\linewidth]{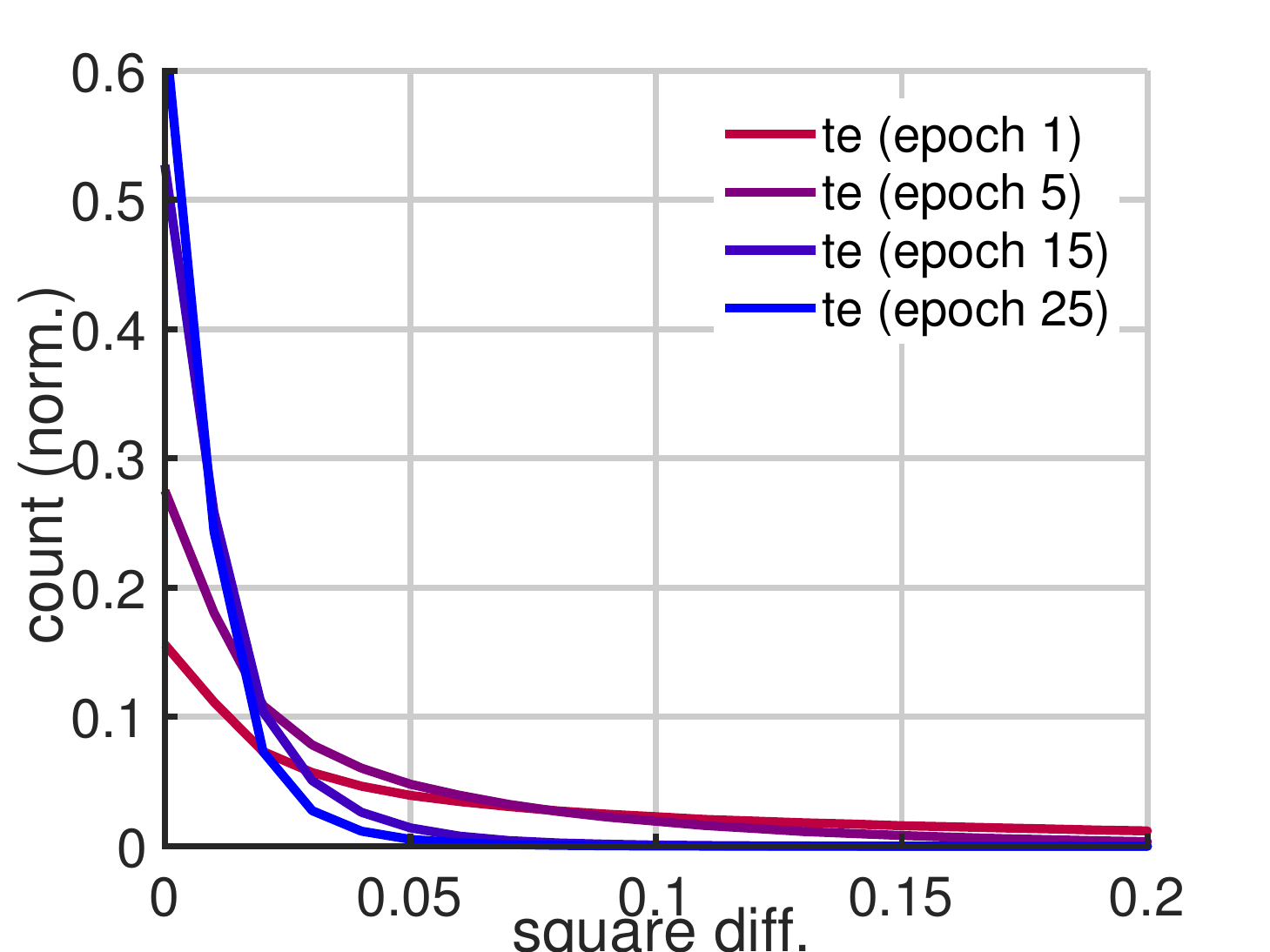}
%\vspace{-0.2cm}
\caption{\label{fig:str12} BoW FC (test)}
%\captionof{subfigure}{bbb\label{fig:strb}}
%\vspace{0.2cm}
\end{subfigure}
\begin{subfigure}[b]{0.245\linewidth}
\centering\includegraphics[trim=0 0 0 0, clip=true,width=0.95\linewidth]{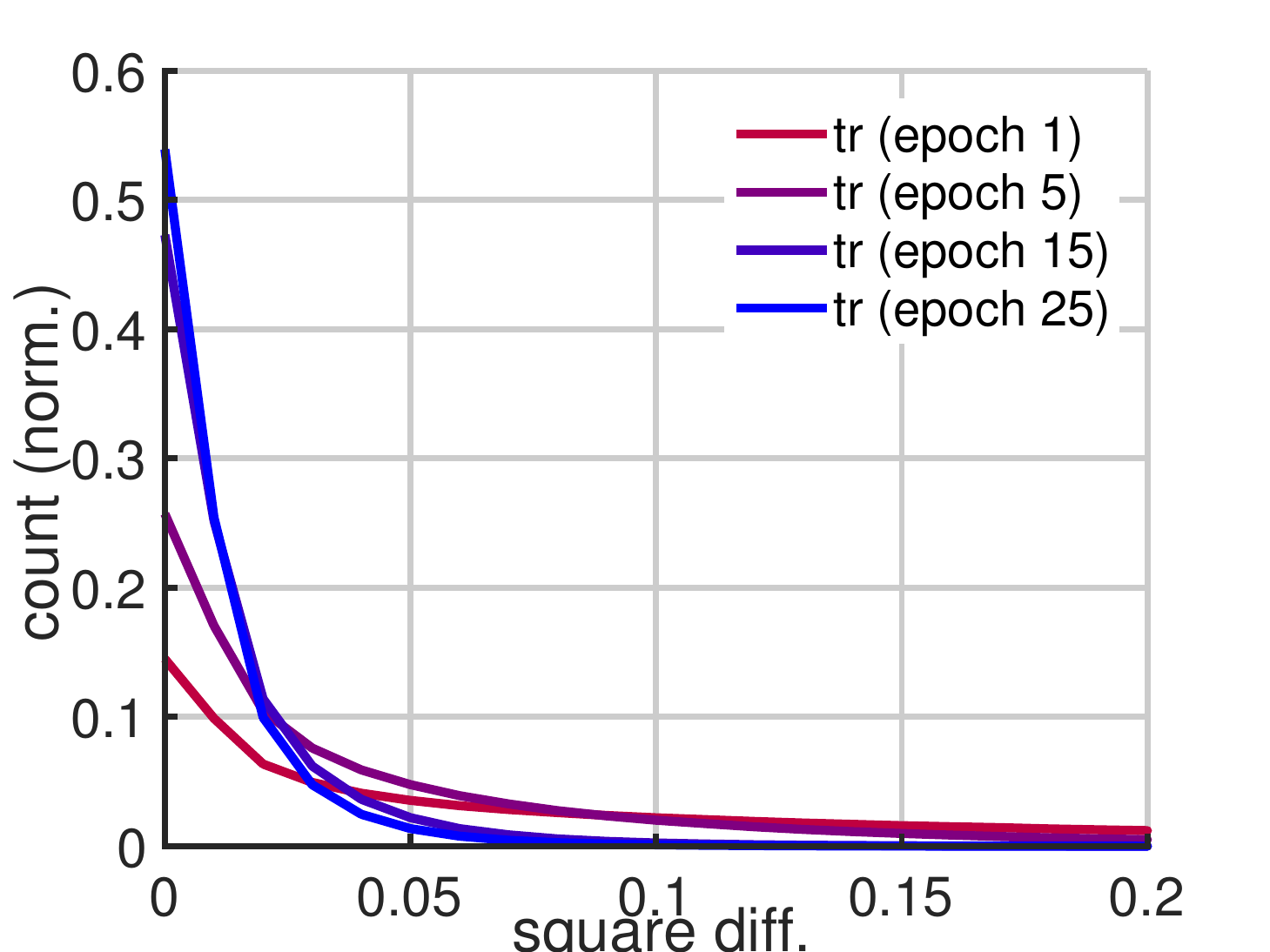}
%\vspace{-0.2cm}
\caption{\label{fig:str13} FV1 FC (train)}
%\captionof{subfigure}{cccc\label{fig:strc}}
%\vspace{0.2cm}
\end{subfigure}
\begin{subfigure}[b]{0.245\linewidth}
\centering\includegraphics[trim=0 0 0 0, clip=true,width=0.95\linewidth]{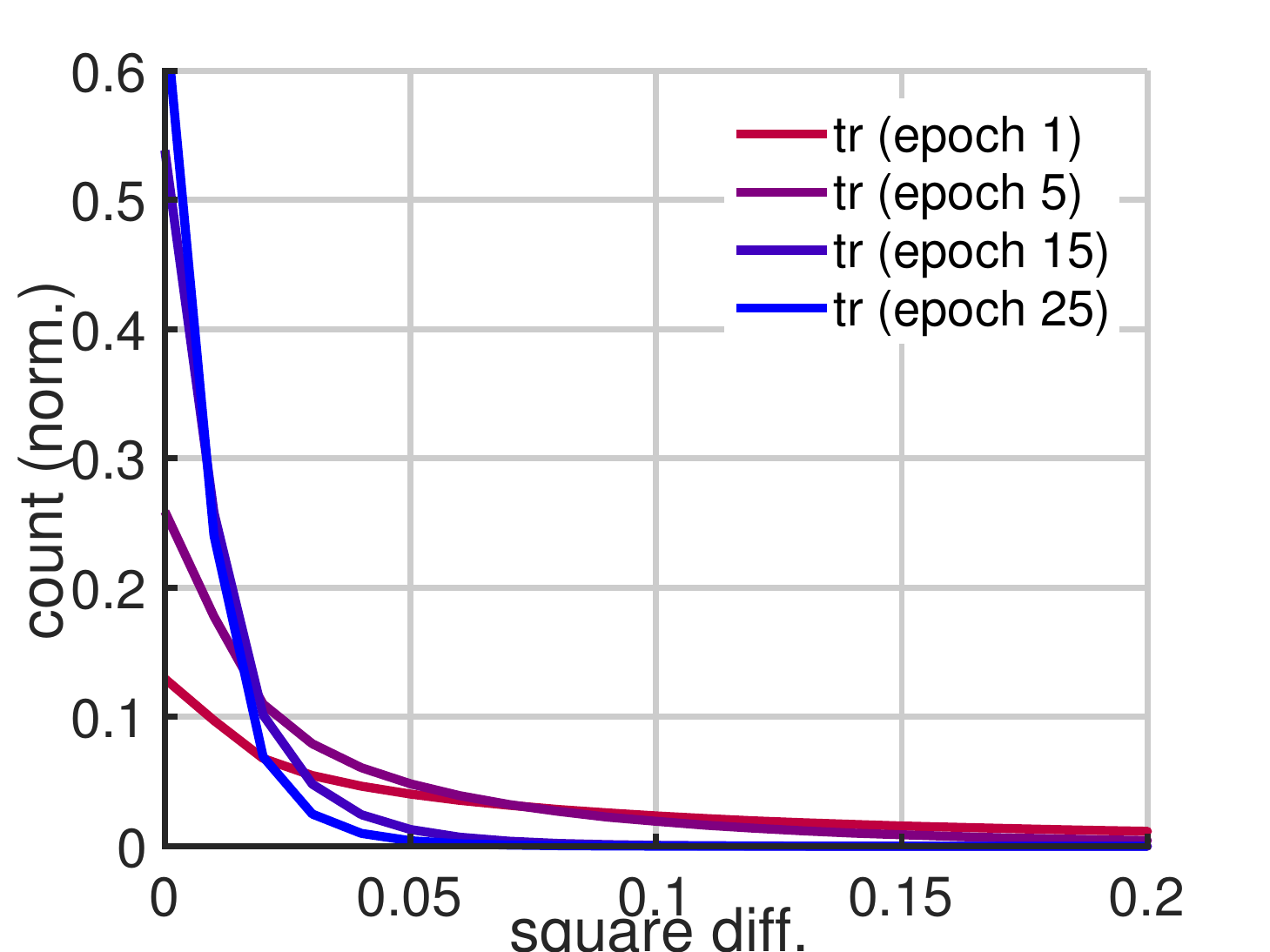}
%\vspace{-0.2cm}
\caption{\label{fig:str15} FV2 FC (train)}
%\captionof{subfigure}{ddd\label{fig:strd}}
%\vspace{0.2cm}
\end{subfigure}
}
\begin{subfigure}[b]{0.245\linewidth}
\centering\includegraphics[trim=0 0 0 0, clip=true,width=0.95\linewidth]{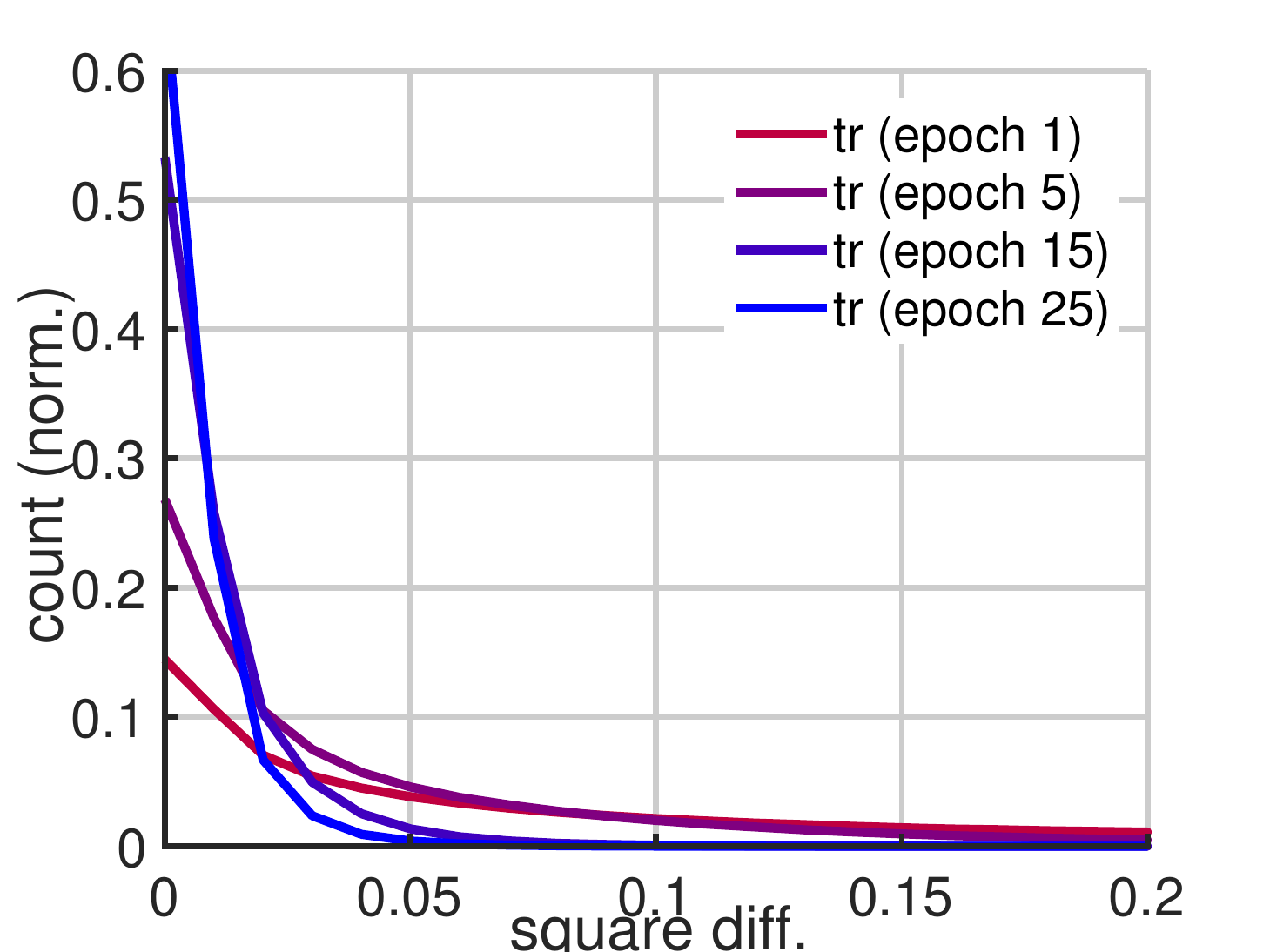}
%\vspace{-0.2cm}
\caption{\label{fig:str21} BoW FC (train)}
%\captionof{subfigure}{aaa\label{fig:stra}}
%\vspace{0.2cm}
\end{subfigure}
\begin{subfigure}[b]{0.245\linewidth}
\centering\includegraphics[trim=0 0 0 0, clip=true,width=0.95\linewidth]{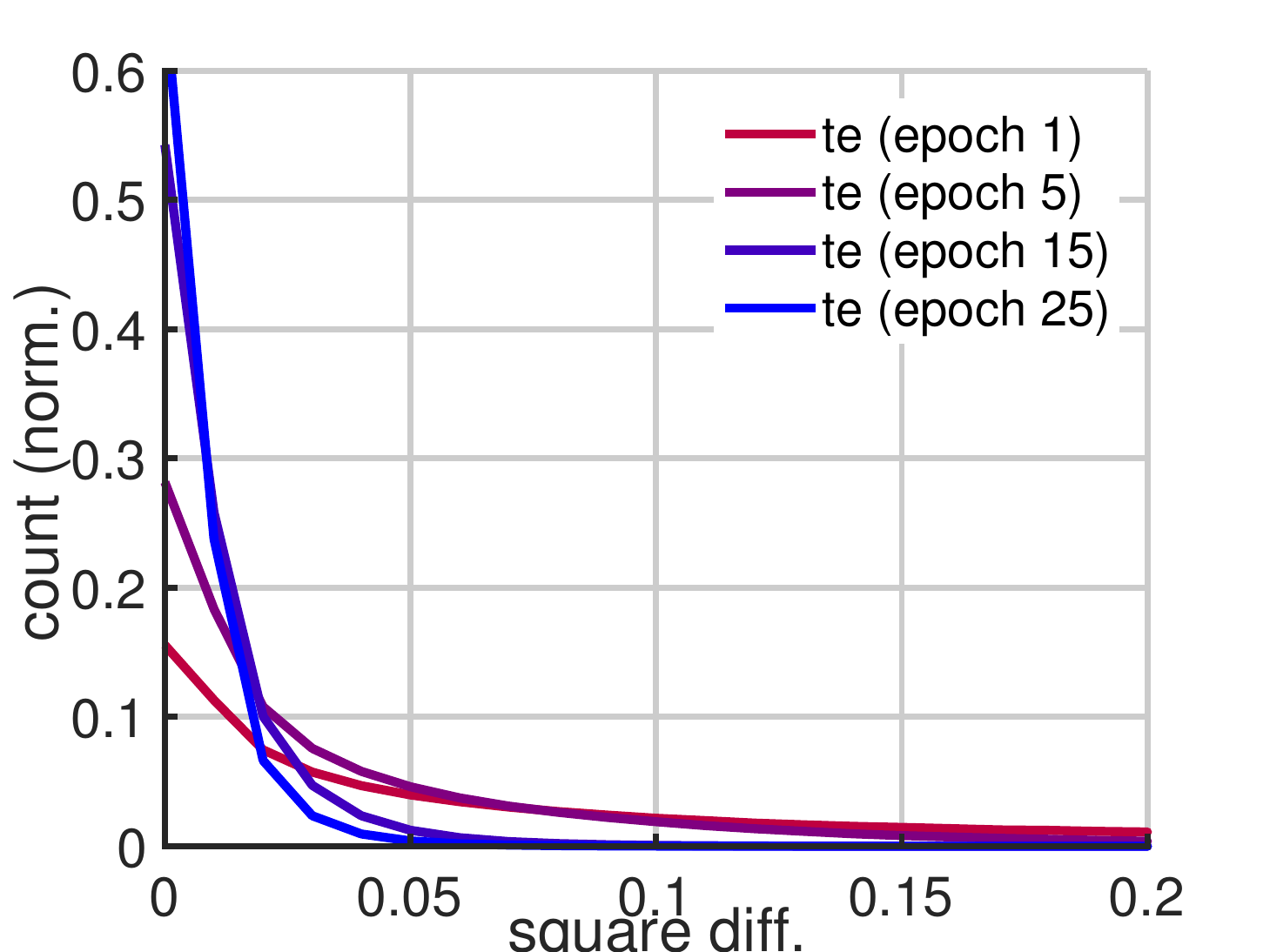}
%\vspace{-0.2cm}
\caption{\label{fig:str22} BoW FC (test)}
%\captionof{subfigure}{bbb\label{fig:strb}}
%\vspace{0.2cm}
\end{subfigure}
\begin{subfigure}[b]{0.245\linewidth}
\centering\includegraphics[trim=0 0 0 0, clip=true,width=0.95\linewidth]{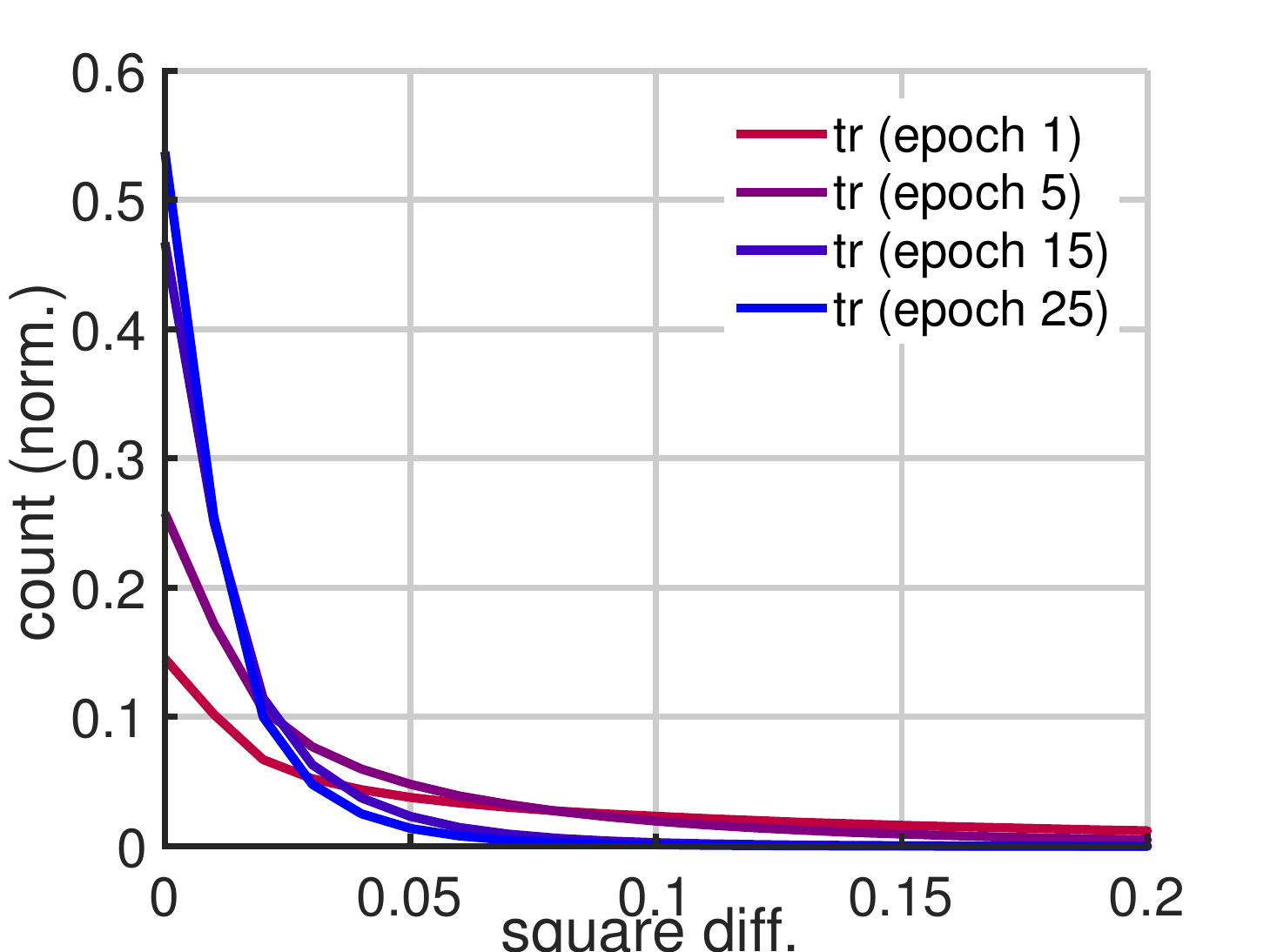}
%\vspace{-0.2cm}
\caption{\label{fig:str23} FV1 FC (train)}
%\captionof{subfigure}{cccc\label{fig:strc}}
%\vspace{0.2cm}
\end{subfigure}
\begin{subfigure}[b]{0.245\linewidth}
\centering\includegraphics[trim=0 0 0 0, clip=true,width=0.95\linewidth]{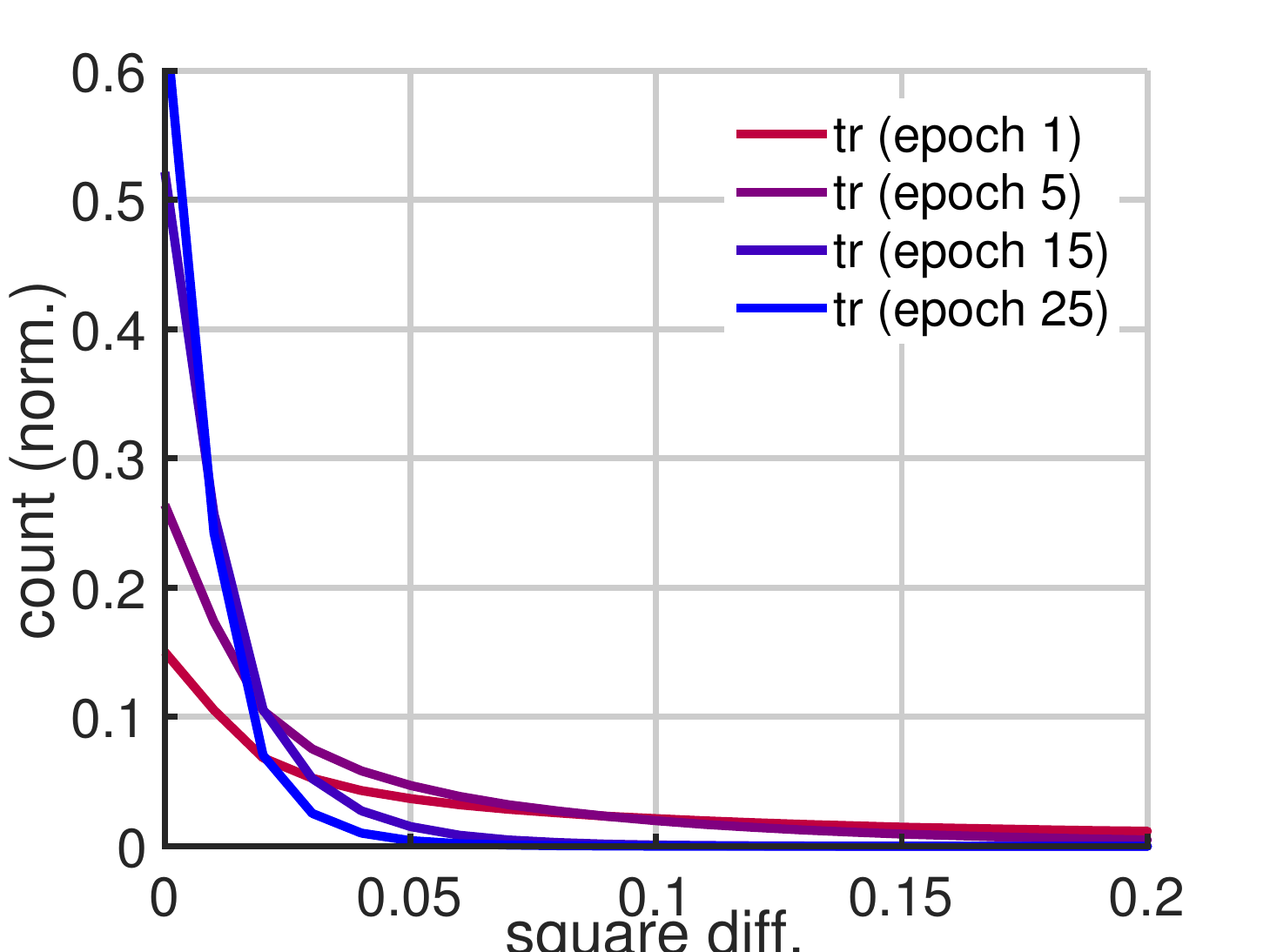}
%\vspace{-0.2cm}
\caption{\label{fig:str25} FV2 FC (train)}
%\captionof{subfigure}{ddd\label{fig:strd}}
%\vspace{0.2cm}
\end{subfigure}
\begin{subfigure}[b]{0.245\linewidth}
\centering\includegraphics[trim=0 0 0 0, clip=true,width=0.95\linewidth]{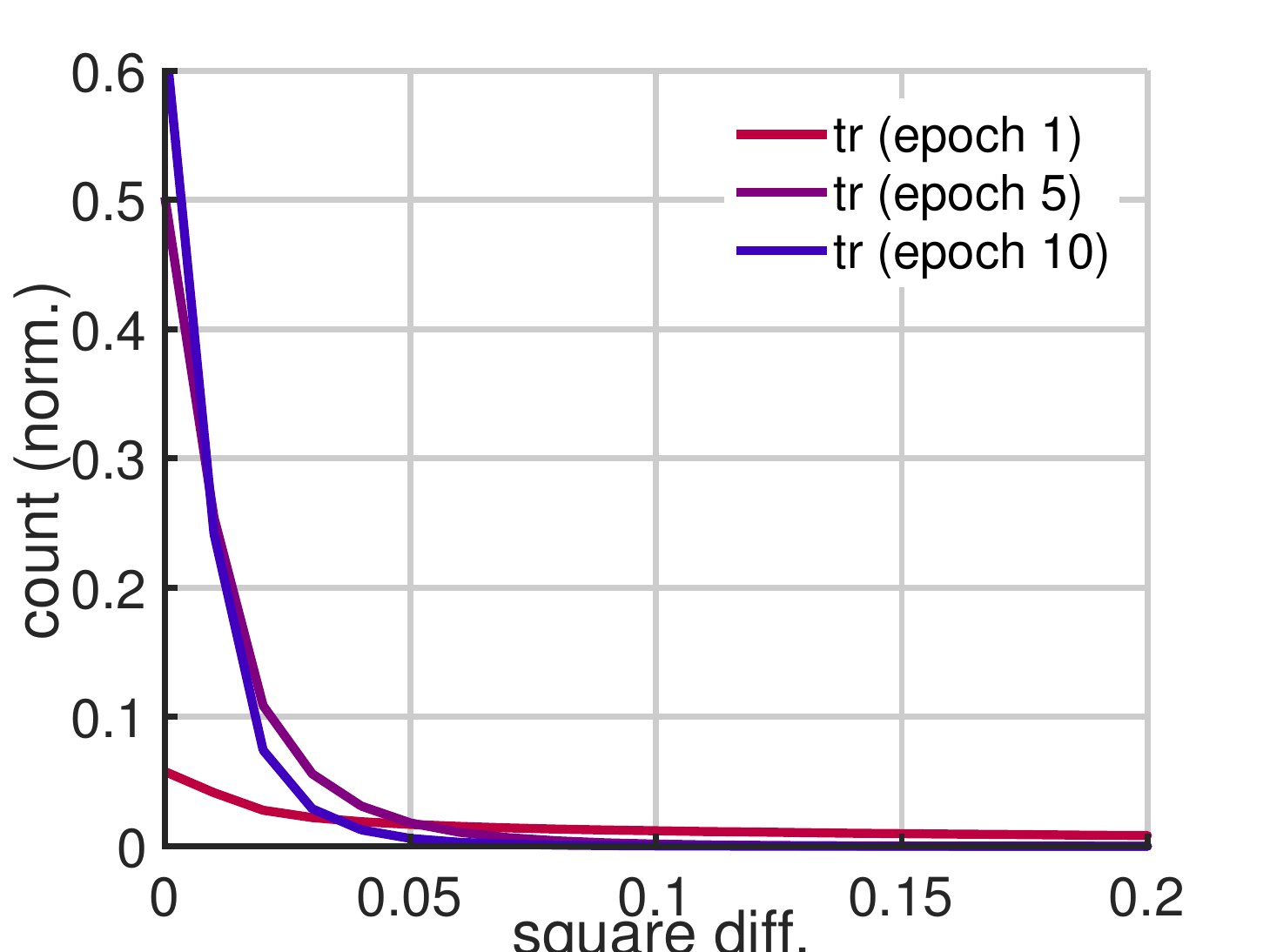}
%\vspace{-0.2cm}
\caption{\label{fig:str31} BoW Conv (train)}
%\captionof{subfigure}{aaa\label{fig:stra}}
%\vspace{0.2cm}
\end{subfigure}
\begin{subfigure}[b]{0.245\linewidth}
\centering\includegraphics[trim=0 0 0 0, clip=true,width=0.95\linewidth]{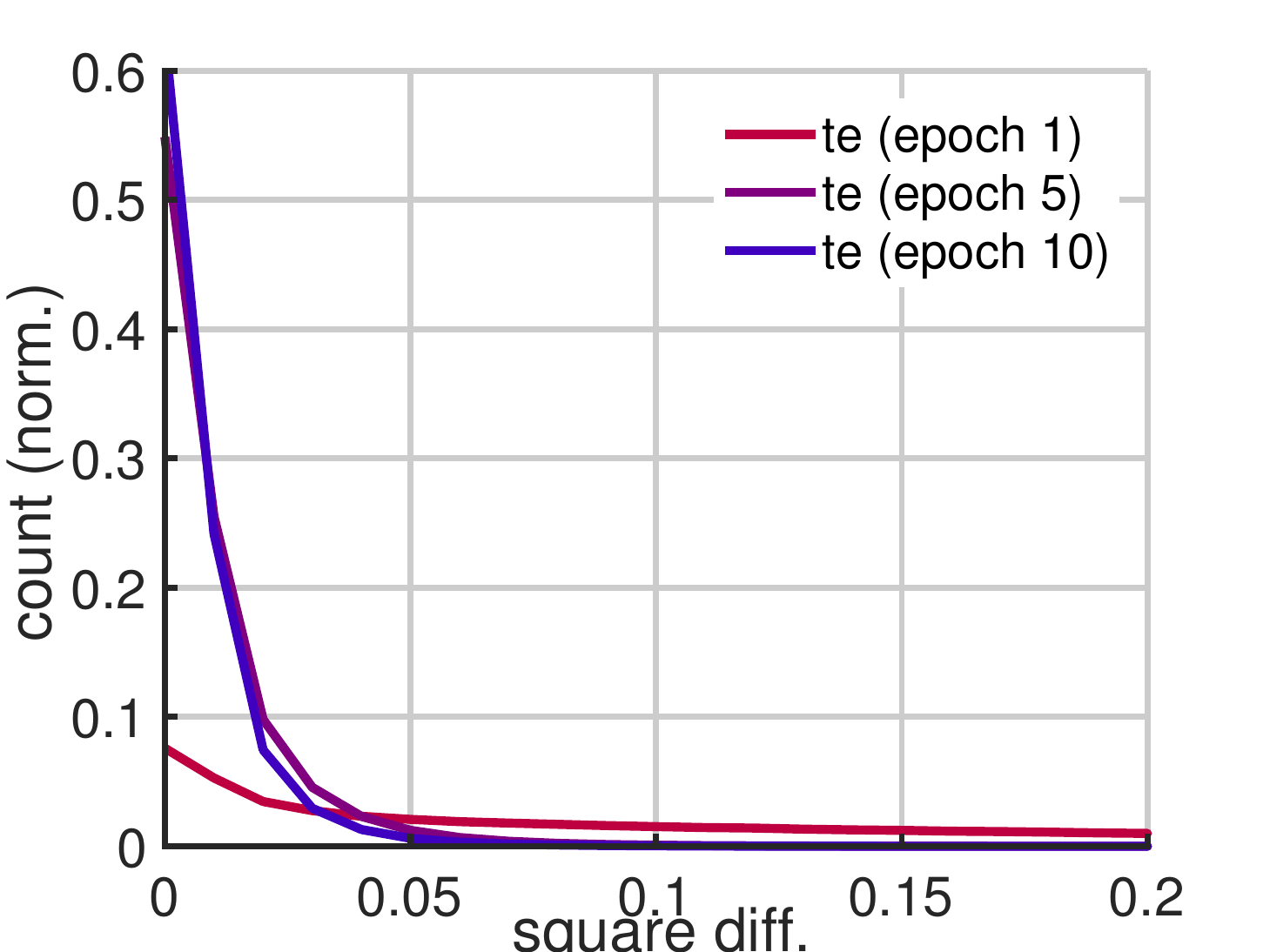}
%\vspace{-0.2cm}
\caption{\label{fig:str32} BoW Conv (test)}
%\captionof{subfigure}{bbb\label{fig:strb}}
%\vspace{0.2cm}
\end{subfigure}
\begin{subfigure}[b]{0.245\linewidth}
\centering\includegraphics[trim=0 0 0 0, clip=true,width=0.95\linewidth]{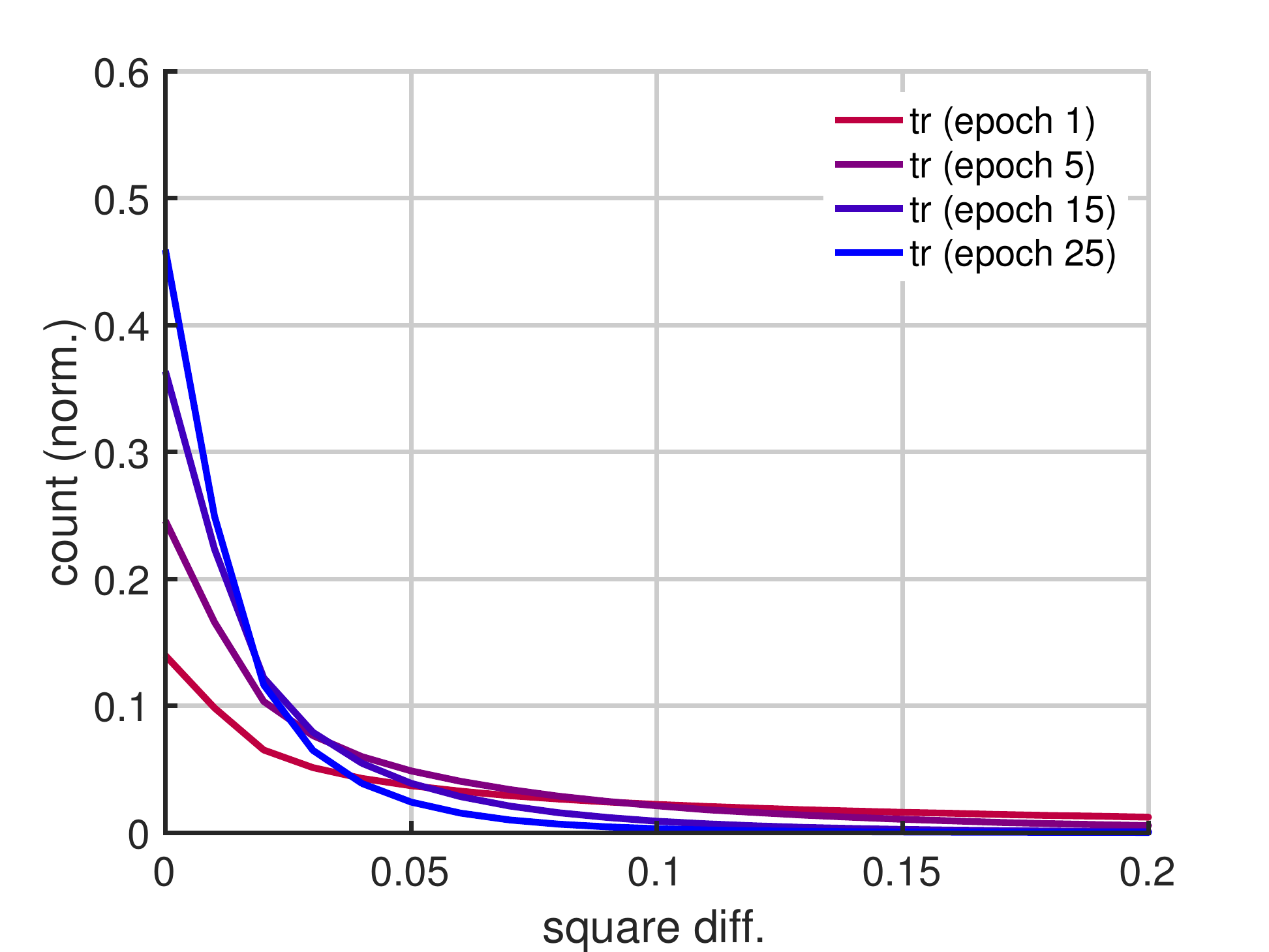}
%\vspace{-0.2cm}
\caption{\label{fig:str33} FV1 FC (train), -SK/PN}
%\captionof{subfigure}{cccc\label{fig:strc}}
%\vspace{0.2cm}
\end{subfigure}
\begin{subfigure}[b]{0.245\linewidth}
\centering\includegraphics[trim=0 0 0 0, clip=true,width=0.95\linewidth]{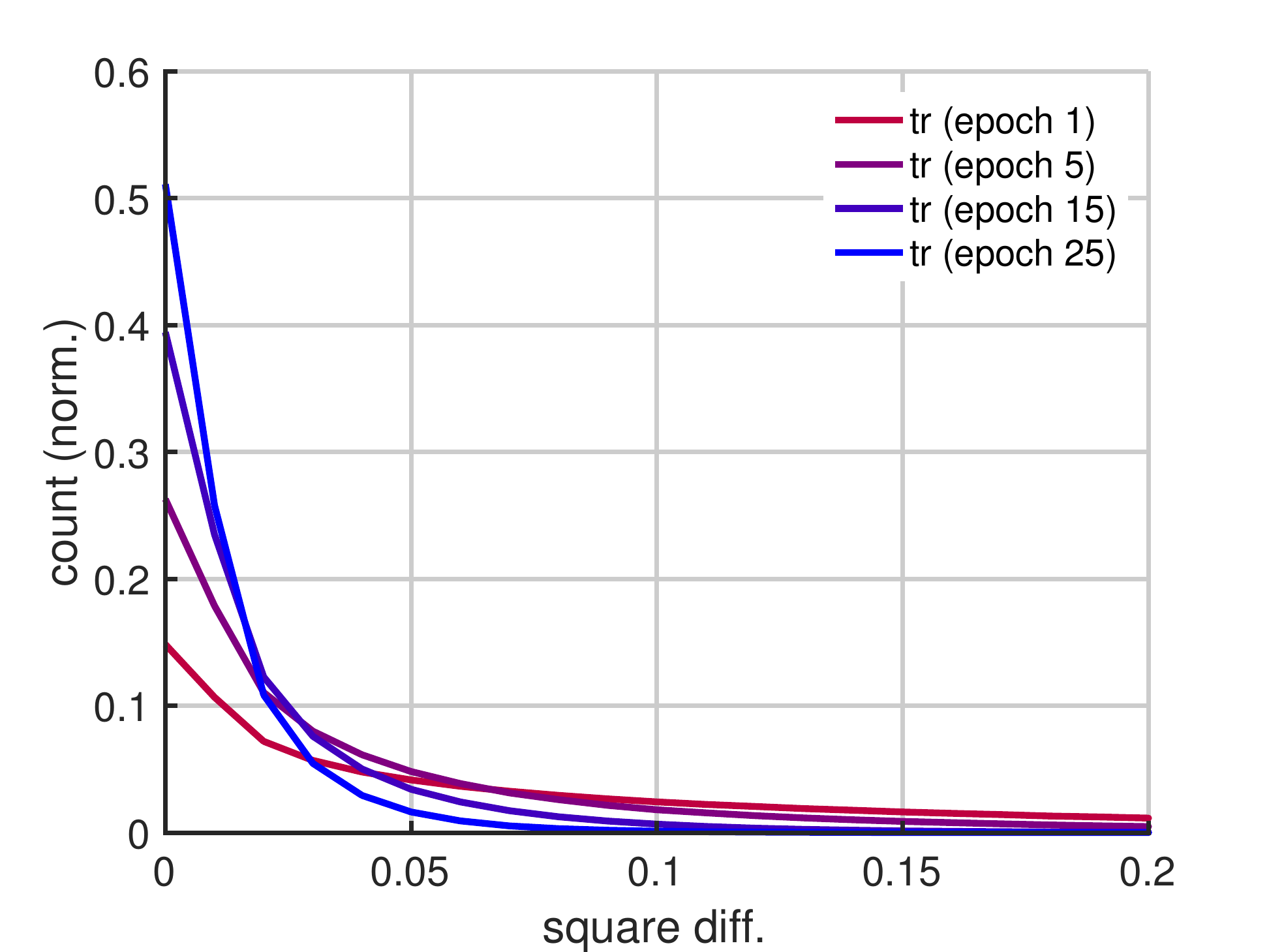}
%\vspace{-0.2cm}
\caption{\label{fig:str35} FV2 FC (train), -SK/PN}
%\captionof{subfigure}{ddd\label{fig:strd}}
%\vspace{0.2cm}
\end{subfigure}
\vspace{-0.3cm}
\caption{Evaluation of the square difference between the hallucinated and ground truth representations on HMDB-51 (split 1). Experiments in the top row use ({\em FC}) streams with sketching and PN. Two leftmost plots in the bottom row use ({\em Conv}) streams. Two rightmost plots in the bottom row use ({\em FC}) streams without sketching/PN ({\em -SK/PN}).}
\vspace{-0.3cm}
%\captionof{figure}{sfsd}
\label{fig:mse}
\end{figure*}
%\end{minipage}

Below we asses (i) the hallucination quality, (ii) provide additional results for higher resolution frames on MPII, and (iii) we provide further details of our pre-processing.

%%%%%%%%% BODY TEXT
\section{Hallucination Quality}
\label{sec:mse}

Below, we provide an analysis of the quality of hallucination of the BoW/FV streams compared to the ground-truth BoW/FV feature vectors. Figure \ref{fig:mse} presents histograms of the square difference between the hallucinated features and ground-truth ones. Specifically, we plot histograms of $\{(\tilde{\psi}_{(bow),mn}-\psi_{(bow),mn})^2, m\!\in\!\idx{1000}, n\!\in\!\mathcal{N}\}$, where index $m$ runs over features $m\!\in\!\idx{1000}$ and $n\!\in\!\mathcal{N}$ runs over each video. Counts for training and testing splits are normalized by 1000 (the number of features) and the number of training and testing videos, respectively. The histograms are computed over bins of size $0.01$ thus allowing us to simply plot continuously looking lines instead of bins.

%\emptybox[8cm]{9.0cm}

Figure \ref{fig:str21} shows that the BoW ground-truth descriptors for the training split are learnt closely by our BoW hallucinating unit based on FC layers ({\em FC}). We capture histograms for epochs $1,5,15,25$ in colors interpolated from red to blue. As one can see, in early epochs, the peak around the first bin is small. As the epochs progress, the peak around the first bin becomes prominent while further bins decrease in size. This indicates that as the training epochs progress, the approximation error becomes smaller and smaller.

Figure \ref{fig:str22} shows that the BoW ground-truth descriptors for the testing split are also approximated closely by the hallucinated BoW descriptors.

We compared histograms for testing and training slits for BoW, first- and second-order FV and observed small differences only. Such a comparison can be conducted by computing the ratio of testing to training bins and it reveals variations between $0.8\times$ and $1.25\times$. Thus, without the loss of clarity, we skip showing plots for FV testing splits.

Figures \ref{fig:str23} and \ref{fig:str25} show that the first- and second-order FV terms ({\em FV1}) and ({\em FV2}) can be also learnt closely by our hallucinating units. We show only the quality of approximation on the training split as behavior on testing splits matches closely the behavior on training splits.

Figures \ref{fig:str31}, \ref{fig:str32}, \ref{fig:str33} and \ref{fig:str35} show  the similar learning/approximation trend for BoW training and testing splits, and the first- and second-order FV terms (training only)  given our hallucinating unit based on FC layers ({\em FC}) with no sketching or PN ({\em -SK/PN}). %However, it appears that ({\em Conv}) learns slower to hallucinate FV descriptors if compared to BoW descriptors. We expect that our ({\em Conv}) unit has not enough learning capacity to fully approximate these representations.

\section{Higher Resolution Frames on MPII}
\label{sec:mpii_bb}

For human-centric pre-processing on MPII denoted by ({\em*}), we observed that the bounding boxes used for extraction of the human subject are of low resolution. Thus, we decided to firstly resize RGB frames to 512 pixels (height) rather than 256 pixels and then compute the corresponding optical flow, and perform extraction of human subjects for which the resolution thus increased $2\!\times$.

The ({\em HAF*+BoW halluc.}), our pipeline with the BoW stream, and  ({\em HAF*+BoW hal.+MSK/PN}) with multiple sketches and PN are computed for the standard 256 pixels (height) denoted by ({\em*}) are given in Table \ref{tab:mpiif2}. 

The ({\em HAF$^\bullet$+BoW halluc.}), our pipeline with the BoW stream, and  ({\em HAF$^\bullet$+BoW hal.+MSK/PN}) pipeline are analogous pipelines but computed for the increased 512 pixel resolution (height) denoted by ({\em$^\bullet$}). According to the table, increasing the resolution $2\!\times$ prior to human detection, extracting subjects in higher resolution and  scaling (to the 256 size for shorter side) yields 1.3\% improvement in accuracy.

\begin{table}[t]%htbp % left bottom right top
\parbox{.99\linewidth}{
\setlength{\tabcolsep}{0.12em}
\renewcommand{\arraystretch}{0.70}
%\fontsize{9}{9}\selectfont
\hspace{-0.3cm}
%\centering
\begin{tabular}{ c | c | c | c | c | c | c | c | c }
 & {\em sp1} & {\em sp2} & {\em sp3} & {\em sp4} & {\em sp5} & {\em sp6} & {\em sp7} & mAP \\
\hline
\kern-0.5em {\fontsize{8}{9}\selectfont HAF*+BoW halluc.}      		 & $78.8$ & $75.0$ & $84.1$ & $76.0$ & $77.0$ & $78.3$ & $75.2$ & $77.8\%$\\
\kern-0.5em {\fontsize{7.5}{9}\selectfont HAF*+BoW hal.+MSK/PN}   & $80.1$ & $79.2$ & $84.8$ & $83.9$ & $80.9$ & $78.5$ & $75.5$ & $80.4\%$\\
\hline
\kern-0.5em {\fontsize{8}{9}\selectfont HAF$^\bullet$+BoW halluc.}      		     & $78.8$ & $78.3$ & $84.2$ & $77.4$ & $77.1$ & $78.3$ & $75.2$ & $78.5\%$\\
\kern-0.5em {\fontsize{7.5}{9}\selectfont HAF$^\bullet$+BoW hal.+MSK/PN}   & $80.8$ & $80.9$ & $85.0$ & $83.9$ & $82.0$ & $79.8$ & $79.6$ & $\mathbf{81.7}\%$\\
\hline
\end{tabular}
\vspace{0.02cm}%\\
}
\caption{Evaluations on  MPII. The ({\em HAF*+BoW halluc.}) is our pipeline with the BoW stream, ({\em *}) denotes human-centric pre-processing for 256 pixels (height) while ({\em HAF*+BoW hal.+MSK/PN}) denotes our pipeline with multiple sketches per BoW followed by Power Norm ({\em PN}). By analogy, ({\em$^\bullet$}) denotes human-centric pre-processing for 512 pixels (height).}
\vspace{-0.3cm}
\label{tab:mpiif2}
\end{table}

\section{Data Pre-processing}
\label{sec:preproc}

 For HMDB-51 and YUP++, we use the data augmentation strategy described in the original authors' papers (\eg, random crop of videos, left-right flips on RGB and optical flow frames. For testing, center crop, no flipping are used.

For the MPII dataset with human-centric pre-processing, human detector is used first. Then, we crop  randomly around the bounding box of human subject (we include it). Finally, we allow scaling,  zooming in, and left-right flips. For longer videos, we sample sequences to form a 64-frame sequence. For short videos (less than 64 frames), we loop the sequence many times to fit its length to the expected input length. 
Lastly, we scale the pixel values of RGB and optical flow frames to the range between $-1$ and $1$.
\end{appendices}

{\small
\bibliographystyle{ieee_fullname}
\bibliography{egbib}
}

\end{document}